\PassOptionsToPackage{table}{xcolor}
\documentclass[10pt,logo]{hitsztechreport}
\usepackage[numbers,sort&compress]{natbib}
\usepackage{multirow,placeins,pgf-pie,wrapfig,float,adjustbox,tabularx,makecell,algorithm,algpseudocode}
\usepackage{array,tikz,subcaption}
\usetikzlibrary{calc,positioning,arrows.meta,shapes.geometric,fit,backgrounds}
\usepackage[noabbrev,nameinlink]{cleveref}
\newcolumntype{Y}{>{\centering\arraybackslash}X}
\newcommand{\paraheading}[1]{\par\addvspace{5pt}\noindent\textbf{#1.}\enspace}
\newcommand{\topparaheading}[1]{\paraheading{#1}}
\newcommand{\modhead}[1]{\paraheading{#1}}

\newcommand{\DIFadd}[1]{#1}\newcommand{\DIFdel}[1]{}
\newcommand{\DIFaddbegin}{}\newcommand{\DIFaddend}{}\newcommand{\DIFdelbegin}{}\newcommand{\DIFdelend}{}

\usepackage{fix-cm}

\usepackage{flafter}

\makeatletter
\renewcommand\AB@affilsepx{\protect\\\protect\Affilfont}
\makeatother
\renewcommand\Affilfont{\normalfont\fontsize{9}{11}\selectfont}
\title{From Actions to Reactions:\\Learning Instrument-Conditioned Tissue Geometry\\for Surgical World Modeling}
\author[1]{Haoran~Shi}
\author[1]{Zhixin~Wu}
\author[2]{Chenxi~Zheng}
\author[3]{Cheng~Xu}
\author[1]{\\Qiang~Huang}
\author[3]{Shengfeng~He}
\author[1,\ensuremath{\dagger}]{Yangyang~Xu}
\author[1]{Jun~Yu}
\affil[1]{Harbin Institute of Technology, Shenzhen}
\affil[2]{South China University of Technology}
\affil[3]{Singapore Management University}
\renewcommand{\reportshorttitle}{From Actions to Reactions}
\paperurl{}
\correspondingauthor{Yangyang Xu}
\makeatletter
\fancypagestyle{firststyle}{%
\fancyhf{}
\fancyhead[L]{\includegraphics[height=14mm]{\institutionlogo}}
\fancyhead[R]{}

}
\let\reportoriginalmaketitle\maketitle
\renewcommand{\maketitle}{%
  \reportoriginalmaketitle
  \vspace{-15pt}% compensate for the source class's trailing title space
  {\noindent\small\textsuperscript{\ensuremath{\dagger}}Corresponding author: Yangyang Xu.\par}
  {\noindent\color{HITBlue}\small\ttfamily\href{https://GenerativeIntelligence-HITSZ.github.io/SurgReact/}{https://GenerativeIntelligence-HITSZ.github.io/SurgReact/}\par}
  \vspace{7pt}
}

\renewcommand{\titlefont}{\color{HITBlue}\normalfont\bfseries\fontsize{19}{23}\selectfont}
\makeatother

\begin{document}
\maketitle
\par\noindent{\headingfont Abstract}\par\vspace{4pt}
{\small Surgical world models predict future surgical observations from visual history and instrument actions, supporting surgical simulation and robot learning. However, existing approaches often prioritize instrument motion and visual fidelity, leaving the resulting tissue reactions implicit in video synthesis. We present \textbf{SurgReact}, a surgical world model that explicitly models tissue reactions before generating future video. Our key idea is to predict tissue responses as pointmaps, allowing deformation to be directly learned and supervised before appearance synthesis. SurgReact first forecasts instrument trajectories and transports the observed instrument surface to construct future instrument pointmaps. An action-conditioned tissue reaction predictor then combines future instrument geometry with historical tissue observations, constructs a geometric reference, and learns coordinate refinements to capture how tissue reacts to instrument motion. The resulting pointmaps guide video generation, while teacher--student distillation helps the generator use imperfect predicted geometry. Experiments demonstrate state-of-the-art video quality and instrument--tissue interaction fidelity on the evaluated benchmarks. Furthermore, SurgReact-generated data improves downstream policy performance, demonstrating its value as a source of training experience for robot learning.
\par}

\providecommand{\DIFadd}[1]{#1}
\providecommand{\DIFdel}[1]{}
\renewcommand{\DIFadd}[1]{{\protect\color{blue}#1}}
\renewcommand{\DIFdel}[1]{{\protect\color{red}\small#1}}
\providecommand{\DIFaddbegin}{}
\providecommand{\DIFaddend}{}
\providecommand{\DIFdelbegin}{}
\providecommand{\DIFdelend}{}
\providecommand{\DIFaddFL}[1]{\DIFadd{#1}}
\providecommand{\DIFdelFL}[1]{\DIFdel{#1}}
\renewcommand{\DIFaddFL}[1]{\DIFadd{#1}}
\renewcommand{\DIFdelFL}[1]{\DIFdel{#1}}
\providecommand{\DIFaddbeginFL}{}
\providecommand{\DIFaddendFL}{}
\providecommand{\DIFdelbeginFL}{}
\providecommand{\DIFdelendFL}{}

\section{Introduction}
\label{sec:intro}

\begin{wrapfigure}[16]{r}{.48\textwidth}
    \centering\captionsetup{font=footnotesize,skip=4pt}
    \includegraphics[width=\linewidth,clip]{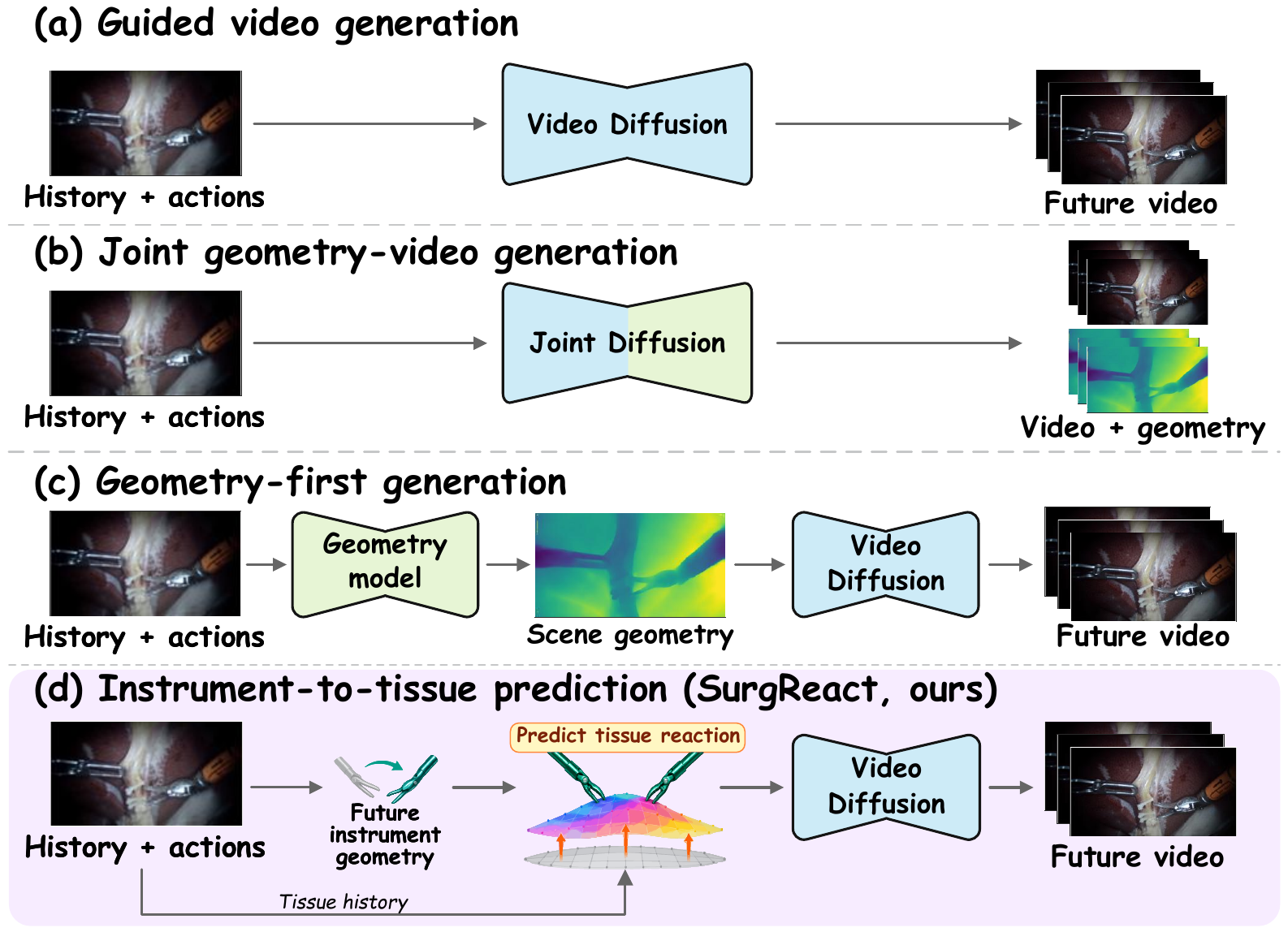}
    
    \caption{\textbf{Representative paradigms in action-conditioned world modeling.}}
                        
    \label{fig:teaser_paradigms}
    
\end{wrapfigure}
Surgical videos provide an important resource for surgical education~\cite{dantas2025video}, visual understanding~\cite{7519080,nwoye2022rendezvous}, and robot learning~\cite{doi:10.1126/scirobotics.adt5254}.
However, collecting expert demonstrations with synchronized robot data is costly~\cite{nelson2026open}.
Action-conditioned video generation offers a scalable way to anticipate the visual consequences of candidate operations. Given surgical observations and prospective robot actions, a surgical world model predicts the corresponding future video.

Recent surgical world models strengthen action control through structured motion guidance for video generation (Fig.~\ref{fig:teaser_paradigms}(a)).
SurgSora~\cite{chen2025surgsora} introduces object-aware motion guidance, while KVLR~\cite{li2026articulated} lifts articulated kinematics into spatial instrument controls and SurgCast~\cite{liu2026surgcast} forecasts future instrument skeletons from robot actions.
SurgVista~\cite{pan2026surgvista} further supervises instrument and tissue trajectories to improve interaction consistency.

Geometric representations further connect robot control to scene evolution.
RoboStereo~\cite{zhang2026robostereo} jointly predicts video and pointmaps (Fig.~\ref{fig:teaser_paradigms}(b)).
In geometry-first generation (Fig.~\ref{fig:teaser_paradigms}(c)), geometry serves as an intermediate representation for subsequent RGB synthesis.
CausalWM~\cite{xu2026causalwm} predicts motion and scene geometry before RGB synthesis, while Kinema4D~\cite{Xu_2026_CVPR} uses kinematically derived robot pointmaps to guide the video generation.

In surgical world modeling, instrument-tissue interaction regions are central to surgical prediction, as they capture the tissue changes induced by surgical manipulation.
However, existing approaches learn tissue reactions implicitly within overall scene prediction, with some methods further emphasizing interaction regions in video supervision.
These strategies constrain the predicted outcome without explicitly modeling how instrument motion produces tissue deformation.
Robot actions directly control the instruments, whereas tissue deformation depends on the resulting contact, the observed tissue shape, and interaction history.
This motivates a finer decomposition within geometry prediction: first establish future instrument geometry, then predict the tissue reaction conditioned on that geometry and tissue history.

We introduce \emph{SurgReact}, a surgical world model that follows this instrument-to-tissue prediction process.
Given visual history and planned actions, SurgReact first forecasts instrument keypoint trajectories and transports the observed instrument surface to construct future instrument pointmaps.
An instrument-conditioned tissue-reaction predictor then integrates these pointmaps with historical tissue geometry and the action sequence, selects reference coordinates, and refines them to produce future scene pointmaps.
These pointmaps guide video generation, while RGB history supplies appearance context.
To help the generator use imperfect predicted geometry, we distill guidance from a teacher conditioned on reference pointmaps into a student conditioned on predicted pointmaps.

Experiments on phantom and ex vivo surgical data demonstrate state-of-the-art instrument-tissue interaction prediction and competitive visual quality across both short- and long-horizon prediction settings. Beyond video prediction, we show that our generated observations support downstream robot learning.

Our contributions are threefold:
\begin{itemize}
    \item We introduce \textbf{SurgReact}, a surgical world model that decomposes geometric prediction into future instrument geometry and instrument-conditioned tissue reactions before RGB synthesis.
    \item We develop an instrument-conditioned tissue-reaction predictor that captures deformation through reference selection and coordinate refinement, and adapt video generation to imperfect predicted geometry through teacher-student distillation from a teacher conditioned on reference pointmaps.
    \item We demonstrate state-of-the-art instrument-tissue interaction prediction and competitive video quality, together with improved downstream surgical action prediction using generated data.
\end{itemize}

\section{Related Work}
\label{sec:related_work}

\topparaheading{Surgical World Modeling}
Surgical world models provide repeatable simulation environments for developing and evaluating robotic surgical behavior, reducing reliance on costly physical trials.
To improve the reliability of simulated operations, recent work has focused on strengthening the links between control inputs, instrument motion, and tissue response.
Spatially conditioned generation methods such as SurgSora~\cite{chen2025surgsora} combine object-aware RGB-D features with trajectory-derived optical flow to control both instrument and tissue motion.
Action-conditioned simulation further connects generated scenes to robot commands. Cosmos-H-Surgical-Simulator~\cite{nelson2026open} learns from synchronized surgical video and kinematics, while KVLR~\cite{li2026articulated} and SurgCast~\cite{liu2026surgcast} explicitly represent instrument motion using projected kinematic control signals and predicted skeletons, respectively.
SurgVista~\cite{pan2026surgvista} introduces cross-frame interaction supervision in the control mechanisms. These advances improve action following and interaction consistency through structured controls and motion supervision.
However, constraining interaction consistency differs from explicitly predicting the geometric tissue response to instrument motion.

\paraheading{Generative Simulation for Robot Learning}
Generative simulation provides a scalable source of interaction experience for robot learning~\cite{pmlr-v305-jang25a}, allowing candidate behaviors to be explored without a physical trial for every action.
Early work such as UniSim~\cite{ICLR2024_c4d66eae} demonstrated the feasibility of using generated interactions to train policies.
Subsequent research has developed the capabilities needed to turn visual prediction into interactive environments.
IRASim~\cite{zhu2025irasim} improves responsiveness to candidate control inputs through fine-grained action-frame alignment. Ctrl-World~\cite{guo2026ctrl} combines multi-view prediction with pose-conditioned memory, while BWM~\cite{bwm2026bwm} maintains a rolling visual history alongside guidance from the initial scene. These works primarily target general robotic manipulation, extending them to surgery requires capturing fine-grained tissue reactions to instrument actions, particularly contact-induced local deformation.

\paraheading{Geometry-Based Modeling}
Explicit geometric prediction supports contact-rich simulation and robot planning~\cite{li2019learning,shi2022robocraft}.
ParticleFormer~\cite{pmlr-v305-huang25c}, PointWorld~\cite{Huang_2026_CVPR}, and 3DPWM~\cite{peri20263d} advance action-conditioned dynamics prediction through multi-material interaction modeling, 3D point flows, and point-cloud completion, respectively.
Pointmaps connect geometric prediction to video synthesis by arranging 3D coordinates on an image grid~\cite{wang2024dust3r}, providing spatially aligned guidance for RGB generation.
RoboStereo~\cite{zhang2026robostereo} couples video and pointmap generation through bidirectional cross-modal interaction, while Kinema4D~\cite{Xu_2026_CVPR} uses kinematically derived robot pointmaps to condition joint RGB--pointmap generation. In contrast, SurgReact predicts instrument-conditioned tissue geometry before video synthesis, making tissue \emph{reaction} an explicit geometric intermediate.

\section{Method}
\label{sec:method}

\begin{figure}[!t]
    \centering
    \includegraphics[width=\textwidth]{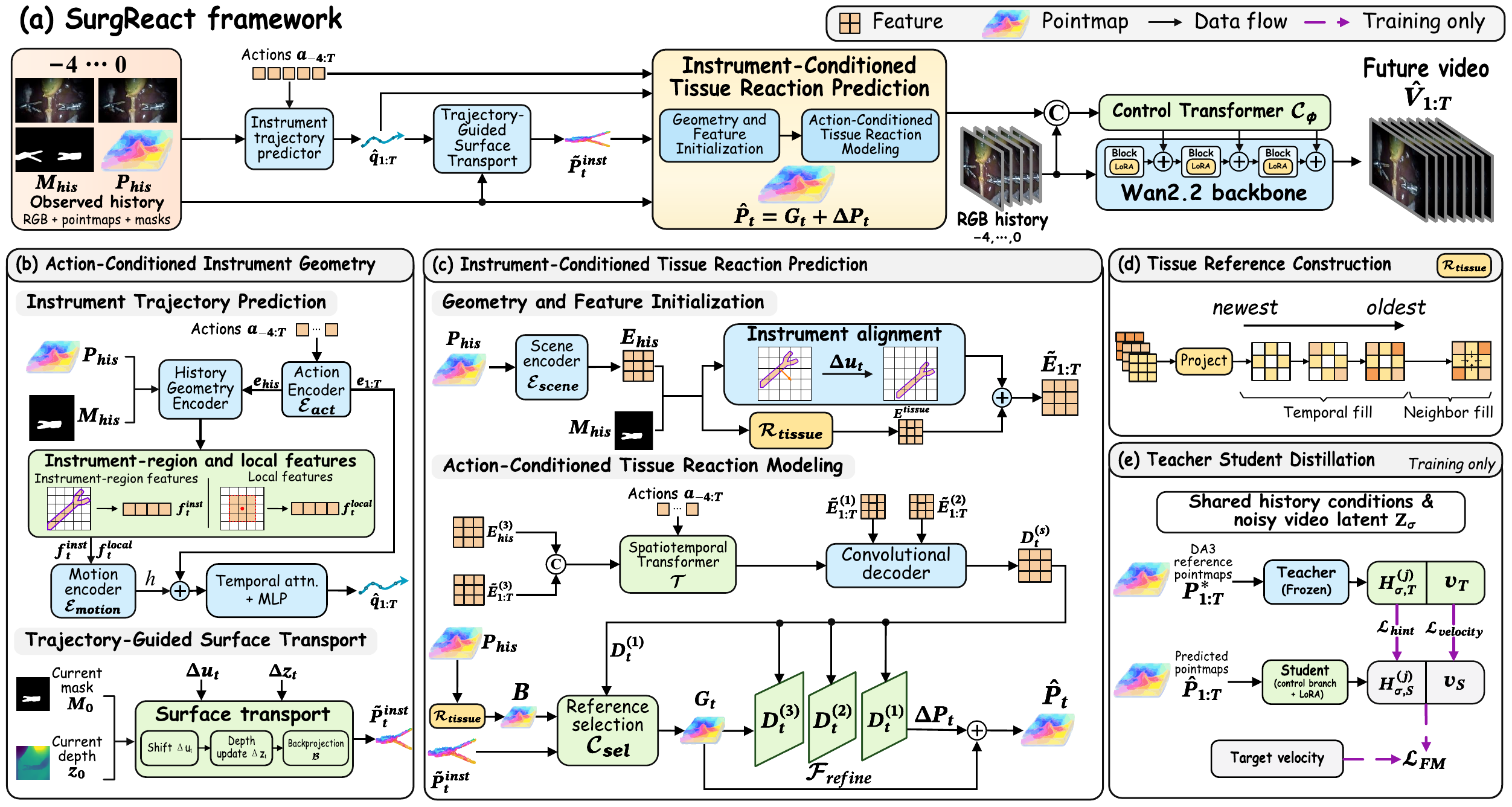}
    
\caption{\textbf{Overview of SurgReact.} (a) SurgReact predicts instrument geometry from actions, then predicts tissue reactions conditioned on this geometry. Future pointmaps guide video generation, while RGB history provides appearance context. (b) Predicted trajectories guide surface transport to construct instrument pointmaps. (c) Instrument and history tissue geometry initialize features for tissue reaction prediction through reference selection and coordinate refinement. (d) Temporal and neighbor filling construct the tissue reference and aligned features. (e) A teacher using reference pointmaps supervises a student using predicted pointmaps through control-feature and velocity alignment to reduce sensitivity to geometry prediction errors.}
    \label{fig:overview}
    
\end{figure}

\subsection{Overview}
\label{sec:overview}

Given five history RGB frames $V_{\mathrm{his}}$ and instrument actions $a_{-4:T}$, SurgReact predicts future frames $\widehat V_{1:T}$, with $t=0$ denoting the current frame. {When predicting RGB frames directly, the model must infer instrument motion and tissue deformation while generating the scene's appearance. To make tissue reactions an explicit prediction target, we first predict image-aligned 3D pointmaps and directly supervise surface changes before RGB generation.} Their image-grid structure then provides spatially aligned guidance for generating the visual appearance of the predicted reactions.

As shown in Fig.~\ref{fig:overview}, the method follows three steps. First, we construct future instrument geometry $\widetilde P_{1:T}^{\mathrm{inst}}$ by predicting instrument
trajectories from actions and observation history. Second, we combine this future instrument geometry with history tissue geometry and actions to predict tissue reactions through reference selection and coordinate refinement, yielding full-scene pointmaps $\widehat P_{1:T}$. Finally, these pointmaps guide video generation, while RGB history supplies appearance context. The geometry predictor thus learns how the scene responds to instrument actions, {and the video generator generates the corresponding visual appearance.}

\subsection{Action-Conditioned Instrument Geometry}
\label{sec:instrument_prediction}

Actions specify planned instrument control, so we first translate them into future instrument geometry as a spatial condition for tissue reaction prediction. Directly forecasting dense instrument pointmaps is challenging because it requires estimating both motion and surface geometry. Since the current surface is already observed, we instead predict a keypoint trajectory and use it to transport that surface. This reuses the observed shape while learning where the instrument will move.

We obtain history pointmaps $P_{\mathrm{his}}$ using pretrained Depth Anything 3 (DA3)~\cite{ICLR2026_e4cd5012} and instrument masks $M_{\mathrm{his}}$ using SAM 3.1~\cite{ICLR2026_e0982cbc} and SurgSLOT~\cite{liu2025surgslot}. Each $P_t\in\mathbb R^{H\times W\times3}$ stores image-aligned 3D coordinates of visible surfaces, while $M_t$ identifies the instrument region. All pointmaps use the current camera coordinate system and a shared normalization, making coordinates comparable across frames.

\noindent\textbf{Instrument Trajectory Prediction.}
We predict a keypoint trajectory for each instrument, and omit the instrument index in the following description for clarity. We select an instrument keypoint based on the geometry of the current mask $M_0$ and track it across the history video using CoTracker3~\cite{Karaev_2025_ICCV}. Its position is $q_t=[u_t;x_t]\in\mathbb R^5$, where $u_t$ is the image-plane position and $x_t$ is the corresponding 3D position from $P_t$.

As shown in Fig.~\ref{fig:overview}(b), the action encoder $\mathcal E_{\mathrm{act}}$ maps $a_{-4:T}$ to time-aware embeddings $\mathbf e_{-4:T}$. To complement the coordinate history with geometric context, we extract instrument-region features $f_t^{\mathrm{inst}}$ using $M_t$ and local features $f_t^{\mathrm{local}}$ around $u_t$. The instrument features summarize the visible instrument, while the local neighborhood can include nearby tissue outside the mask. The motion encoder $\mathcal E_{\mathrm{motion}}$ summarizes both features, keypoint positions, and displacements relative to the current position to produce the motion history representation $h$:
\begin{equation}
\label{eq:motion_history}
h=\mathcal E_{\mathrm{motion}}\!\left(
\bigl(
f_t^{\mathrm{local}},
f_t^{\mathrm{inst}},
q_t,q_t-q_0
\bigr)_{t=-4}^{0}
\right).
\end{equation}

We combine $h$ with future action embeddings and use temporal attention followed by a regression head, denoted by $\mathcal H$, to predict offsets from the current keypoint:
\begin{equation}
\label{eq:instrument_position}
\Delta q_{1:T}=\mathcal H\!\left((h+\mathbf e_t)_{t=1}^{T}\right),
\quad
\widehat q_t=q_0+\Delta q_t.
\end{equation}

We train these networks with $\mathcal L_{\mathrm{pos}}$, which minimizes the mean per-coordinate $\ell_1$ errors between predicted and pseudo-label trajectories in image and 3D space over future frames. 

\noindent\textbf{Trajectory-Guided Surface Transport.}
The keypoint trajectory gives sparse positions, whereas tissue reaction prediction requires an estimate of the instrument surface at each future step. We construct this estimate by shifting the current mask $M_0$ and its depth by $\Delta u_t=\widehat u_t-u_0$, then adding the predicted keypoint depth change $\Delta z_t$. At each valid pixel $u$ within the shifted mask,
\begin{equation}
\label{eq:instrument_pointmap_construction}
\widetilde P_t^{\mathrm{inst}}(u)
=\mathcal B\!\left(z_0(u-\Delta u_t)+\Delta z_t,\;u\right).
\end{equation}

Here, $z_0$ is recovered from $P_0$ in the original depth units, and $\Delta z_t$ is converted to the same units. The operator $\mathcal B$ backprojects the updated depth using the current camera intrinsics and applies the shared coordinate normalization. {The resulting pointmaps \(\widetilde P_{1:T}^{\mathrm{inst}}\) provide an initial estimate of future instrument geometry.}

\subsection{Instrument-Conditioned Tissue Reaction}
\label{sec:geometry_prediction}

Future instrument geometry specifies where the instrument is expected to move, but does not by itself determine tissue reaction. The response also depends on the observed tissue shape and interaction history. We therefore initialize the scene from predicted instrument placement and history tissue observations, infer their reaction under the action sequence, and convert the resulting features into future surface coordinates through reference selection and residual prediction.

\noindent\textbf{Geometry and Feature Initialization.}
We first prepare spatially aligned geometry and features so that the reaction predictor can relate future instrument placement to the observed tissue. As illustrated in Fig.~\ref{fig:overview}(c), a scene encoder $\mathcal E_{\mathrm{scene}}$ processes the complete history pointmaps, retaining both instrument and tissue geometry:
\begin{equation}
\label{eq:scene_encoding}
\left(
E_{\mathrm{his}}^{(1)},
E_{\mathrm{his}}^{(2)},
E_{\mathrm{his}}^{(3)}
\right)
=\mathcal E_{\mathrm{scene}}(P_{\mathrm{his}}),
\end{equation}
the features $E_{\mathrm{his}}^{(s)}=(E_{-4}^{(s)},\ldots,E_0^{(s)})$, $s=1,2,3$, describe history geometry from fine to coarse. The following feature operations are applied at each scale, with the scale index omitted for clarity.

\emph{Instrument alignment.}
Using the trajectory from Eq.~\eqref{eq:instrument_position}, we shift current features within $M_0$ by $\Delta u_t$, converted to the corresponding feature-grid coordinates. The resulting $E_t^{\mathrm{inst}}$ places the observed instrument features at their predicted future locations.

\emph{Tissue reference construction.}
Current instrument occlusions can hide tissue that was visible earlier. We therefore aggregate history tissue observations into a reference $B\in\mathbb R^{H\times W\times3}$ and aligned features $E^{\mathrm{tissue}}$:
\begin{equation}
\label{eq:tissue_geometry_features}
\left(B,E^{\mathrm{tissue}}\right)
=\mathcal R_{\mathrm{tissue}}
(P_{\mathrm{his}},E_{\mathrm{his}};M_{\mathrm{his}}).
\end{equation}

{As shown in Fig.~\ref{fig:overview}(d), $\mathcal R_{\mathrm{tissue}}$ constructs $B$ and $E^{\mathrm{tissue}}$ through projection, temporal filling, and neighbor filling. We first project history tissue geometry onto the current image grid, excluding instrument regions. Temporal filling~\cite{purica2015improved} proceeds from the newest frame to the oldest, using earlier frames to fill remaining gaps. We gather features from the same source pixels to keep geometry and features aligned. For neighbor filling, we follow the nearest-neighbor depth filling in Chen et al.~\cite{chen2018estimating}. At each remaining gap, we backproject the nearest available tissue depth and copy the feature from the same neighboring pixel.} Both $B$ and $E^{\mathrm{tissue}}$ are shared across future steps: they supply the observed starting state, while future tissue changes remain to be predicted.

\emph{Future feature assembly.}
We assign an initial feature source at each image pixel using the predicted instrument geometry and tissue reference. Where both depths are valid, we choose the nearer surface; otherwise, the shifted instrument mask determines the source. For each feature cell $p$, $\alpha_t(p)$ is the fraction of corresponding image pixels assigned to the instrument. We obtain
\begin{equation}
\label{eq:future_feature_initialization}
\widetilde E_t(p)=\alpha_t(p)E_t^{\mathrm{inst}}(p)+\bigl(1-\alpha_t(p)\bigr)E^{\mathrm{tissue}}(p).
\end{equation}
{The resulting features $\widetilde E_{1:T}$ combine predicted instrument motion with observed tissue geometry. They provide the initial features for modeling tissue reaction in the next step.}

\noindent\textbf{Action-Conditioned Tissue Reaction Modeling.}
The initialized layout must now be updated to reflect how tissue responds to the planned operation. We first infer action-conditioned scene context, then select reference coordinates and predict their changes.

\emph{Spatiotemporal reasoning.}
We jointly process the coarsest history and initialized future features with a spatiotemporal Transformer $\mathcal T$:
\begin{equation}
\label{eq:scene_features}
E_{1:T}^{\mathrm{out}}
=\mathcal T\!\left(
E_{\mathrm{his}}^{(3)},
\widetilde E_{1:T}^{(3)};
a_{-4:T}
\right).
\end{equation}

Following IRASim~\cite{zhu2025irasim}, we use frame-wise AdaLN to condition each frame's features on its corresponding action. {Thus, we obtain action-conditioned future scene features \(E_t^{\mathrm{out}}\) that integrate predicted instrument motion with history tissue geometry.} A convolutional decoder upsamples $E_t^{\mathrm{out}}$ and combines it with finer-scale $\widetilde E_t^{(s)}$, producing multiscale features $D_t^{(s)}$. The full-resolution features $D_t^{(1)}$ support pixel-wise decisions, while the coarser features retain broader context for coordinate refinement.

\emph{Reference selection.}
Both transported instrument geometry and historical tissue geometry may deviate from the future surface, so we learn which source provides a more suitable reference at each pixel. A convolutional selector predicts relative confidences:
\begin{equation}
\label{eq:selection_confidence}
\left(c_t^{\mathrm{inst}},c_t^{\mathrm{tissue}}\right)
=\operatorname{softmax}\!\left(
\mathcal C_{\mathrm{sel}}\!\left(
[D_t^{(1)};\widetilde P_t^{\mathrm{inst}};B]
\right)\right),
\end{equation}
where $[\cdot;\cdot]$ denotes channel concatenation. The geometry reference \(G_t\in\mathbb R^{H\times W\times3}\) is constructed as
\begin{equation}
\label{eq:geometry_reference}
G_t(u)=
\begin{cases}
\widetilde P_t^{\mathrm{inst}}(u),
&c_t^{\mathrm{inst}}(u)>c_t^{\mathrm{tissue}}(u),\\
B(u),&\text{otherwise}.
\end{cases}
\end{equation}

Unlike the earlier feature fusion, this step selects a surface coordinate rather than mixing feature vectors.
During training, DA3 estimates reference pointmaps $P_t^*$ from the future training frames. The candidate with smaller coordinate $\ell_1$ error supplies the reference-source label for cross-entropy loss $\mathcal L_{\mathrm{sel}}$. This supervision favors a starting coordinate that requires a smaller correction.

\emph{Coordinate refinement.}
We predict the remaining changes relative to $G_t$ using the action-conditioned features. The refinement network combines features and reference geometry at matching resolutions:
\begin{equation}
\label{eq:multiscale_residual}
\Delta P_t=\mathcal F_{\mathrm{refine}}\!\left(
D_t^{(3)},D_t^{(2)},D_t^{(1)};G_t
\right).
\end{equation}
Starting from $D_t^{(3)}$, it progressively upsamples features and fuses them with $D_t^{(2)}$ and $D_t^{(1)}$, incorporating $G_t$ at every scale. The resulting three-channel residual corrects the normalized $X$, $Y$, and $Z$ coordinates:
\begin{equation}
\label{eq:geometry_refinement}
\widehat P_t=G_t+\Delta P_t.
\end{equation}

The residual models tissue responses relative to the selected reference and corrects errors in the transported instrument geometry, yielding a future pointmap of the visible scene.

\noindent\textbf{Geometry Supervision.}
We supervise predicted coordinates and surface shape using the reference pointmaps $P_t^*$. The total objective also constrains reference selection and the upstream instrument trajectory:
\begin{equation}
\label{eq:geometry_objective}
\mathcal L_{\mathrm{geo}}
=\mathcal L_{\mathrm{xyz}}
+\lambda_{\mathrm{surf}}\mathcal L_{\mathrm{surf}}
+\lambda_{\mathrm{sel}}\mathcal L_{\mathrm{sel}}
+\lambda_{\mathrm{pos}}\mathcal L_{\mathrm{pos}},
\end{equation}
where $\mathcal L_{\mathrm{xyz}}$ is the coordinate $\ell_1$ loss. Inspired by SurgVista~\cite{pan2026surgvista}, we assign higher weights to interaction regions. Our surface loss \(\mathcal L_{\mathrm{surf}}\) matches 3D coordinate differences between the predicted and reference pointmaps, drawing on depth-gradient supervision in DA3~\cite{ICLR2026_e4cd5012}. The cross-entropy loss $\mathcal L_{\mathrm{sel}}$ supervises geometry-source confidences, and $\mathcal L_{\mathrm{pos}}$ supervises predicted instrument keypoint positions. The $\lambda$ coefficients weight the corresponding losses. Detailed loss definitions and training settings are provided in the supplementary material.

The resulting pointmaps $\widehat P_{1:T}$ make the predicted tissue reactions and instrument motion available as spatial conditions for video generation.

\subsection{Geometry-Conditioned Video Generation}
\label{sec:video_generation}

We generate future video using predicted pointmaps as spatial guidance and RGB history as appearance context. Inspired by ControlNet~\cite{zhang2023controlnet} and Uni3C~\cite{cao2025uni3c}, we use a lightweight control branch to guide a pre-trained video diffusion model.

\noindent\textbf{Geometry Conditioning.}
We use a frozen video VAE to encode history and predicted pointmaps as $\mathbf g$, and RGB history as $\mathbf Z_{\mathrm{his}}$. To adapt geometry guidance to the current video state, the control Transformer $\mathcal C_\phi$ also takes the noisy video latent $\mathbf Z_\sigma$ at noise level $\sigma$:
\begin{equation}
\label{eq:geometry_control}
\left\{\mathbf H_\sigma^{(j)}\right\}_{j=1}^{J}
=\mathcal C_\phi([\mathbf Z_\sigma;\mathbf g],\sigma).
\end{equation}

Here, $J$ is the number of backbone blocks receiving geometry guidance. Each projected control feature $\mathbf H_\sigma^{(j)}$ is added to the output of its corresponding block. We train only the control branch and backbone LoRA~\cite{hu2022lora}.

\noindent\textbf{Teacher Student Distillation.}
To reduce the impact of pointmap prediction errors on video quality, we use a frozen teacher conditioned on reference pointmaps $P_{1:T}^*$ to supervise a student conditioned on predicted pointmaps $\widehat P_{1:T}$~\cite{lopezpaz2016unifying}. 

As shown in Fig.~\ref{fig:overview}(e), both receive the same history conditions, noisy video latent, and noise level. We align their projected control features $\mathbf H_{\sigma,\mathrm T}^{(j)}$ and $\mathbf H_{\sigma,\mathrm S}^{(j)}$ at corresponding backbone blocks with $\mathcal L_{\mathrm{hint}}$~\cite{romero2015fitnets}. We also align their predicted future latent velocities $\mathbf v_{\mathrm T}$ and $\mathbf v_{\mathrm S}$ with $\mathcal L_{\mathrm{velocity}}$, where $\mathrm T$ and $\mathrm S$ denote the teacher and student, respectively. Both losses use squared $\ell_2$ errors. We retain the flow-matching loss $\mathcal L_{\mathrm{FM}}$~\cite{lipman2023flow} for supervision from training videos. The training objective is
\begin{equation}
\label{eq:video_objective}
\mathcal L_{\mathrm{video}}
=
\mathcal L_{\mathrm{FM}}
+\lambda_h\mathcal L_{\mathrm{hint}}
+\lambda_v\mathcal L_{\mathrm{velocity}},
\end{equation}
Here, $\lambda_h$ and $\lambda_v$ weight the distillation losses. Only the student is used at inference. Loss normalization and training details are provided in the supplementary material.

\section{Experiments}
\label{sec:experiments}

\subsection{Experimental Setup}
\label{sec:experimental_setup}

\noindent\textbf{Datasets.}
We use surgical videos and synchronized robot kinematics from Open-H-Embodiment~\cite{nelson2026open}. Our experiments use 29 surgical subsets comprising 79.05 hours of video across 33,846 view-specific sequences from 25,418 episodes, covering diverse tasks and collection environments in phantom and ex vivo settings. We train and evaluate models separately in these two domains. Data are split into training and test sets at the episode level, with all views of each episode assigned to the same split. We construct an ablation dataset from 6 representative task subsets of the main experimental data, covering both phantom and ex vivo settings. It contains 9,403 episodes and is partitioned into training and test sets at an 8:1 ratio. The selected subset provides diverse settings in our ablation study.

\noindent\textbf{Implementation Details.}
We train the geometry predictor and video generator separately for 50K iterations each. The video generator builds on pre-trained Wan2.2 TI2V-5B~\cite{wan2025wan} at a resolution of $512\times288$, optimizing the control branch and backbone LoRA parameters~\cite{hu2022lora}. Given five history frames and the future action sequence, SurgReact predicts 28 future frames for short-horizon generation and 80 for long-horizon generation at 15 fps, corresponding to approximately 1.87 and 5.33 seconds, respectively.

\begin{figure}[tbp]
\centering
\includegraphics[width=\textwidth]{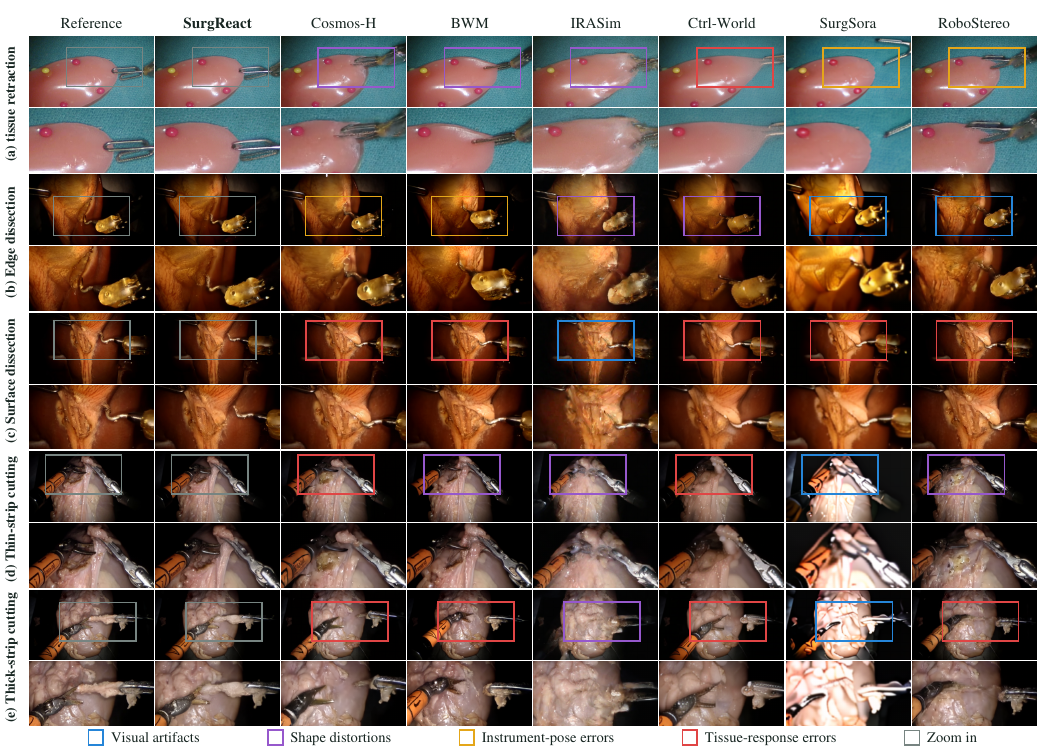}

\captionsetup{singlelinecheck=false}
\caption{\textbf{Qualitative comparison across surgical scenarios.}
Each scenario compares the same future frame across methods, with zoomed-in views below.
Box colors indicate error types, and gray boxes mark regions for visual comparison.}
\label{fig:interaction_comparison}

\end{figure}

\noindent\textbf{Baselines and Evaluation Metrics.}
We compare SurgReact with Cosmos-H~\cite{nelson2026open},
BWM~\cite{bwm2026bwm}, IRASim~\cite{zhu2025irasim},
Ctrl-World~\cite{guo2026ctrl}, SurgSora~\cite{chen2025surgsora},
and RoboStereo~\cite{zhang2026robostereo}.
All methods use the same training split, test windows, and
evaluation protocol.

We assess visual quality using L1, PSNR, SSIM~\cite{1284395},
and LPIPS~\cite{zhang2018unreasonable}.
Following SurgWorld-Bench~\cite{pan2026surgvista}, we report
AJ and $\delta_{\mathrm{avg}}^{\mathrm{vis}}$~\cite{NEURIPS2022_58168e8a}
separately for instrument and tissue trajectories.
We evaluate mean pointwise Euclidean distance in normalized coordinates
($E_{\mathrm{3D}}$) and mean surface-normal angular error in degrees
($E_{\mathrm{n}}$) against DA3 reference geometry over Global, Instrument,
and Interaction regions.

\begin{table}[t]\centering
\caption{\textbf{Quantitative comparison} on instrument-tissue interaction and visual quality in short and long windows. Best and second-best results are bold and underlined, respectively.}

\label{tab:interaction_comparison}
\small
\setlength{\tabcolsep}{3pt}\renewcommand{\arraystretch}{1.02}
\textbf{(a) Short window}\par\smallskip
\begin{adjustbox}{max width=\linewidth}\begin{tabular}{@{}cl*{8}{c}@{}}\toprule
& \multirow{2}{*}{\textbf{Method}} & \multicolumn{2}{c}{\textbf{Instr. Motion}} & \multicolumn{2}{c}{\textbf{Tissue Resp.}} & \multicolumn{4}{c}{\textbf{Visual Quality}}\\
\cmidrule(lr){3-4}\cmidrule(lr){5-6}\cmidrule(l){7-10}
&& AJ$\uparrow$ & $\delta_{\mathrm{avg}}^{\mathrm{vis}}\uparrow$ & AJ$\uparrow$ & $\delta_{\mathrm{avg}}^{\mathrm{vis}}\uparrow$ & L1$\downarrow$ & PSNR$\uparrow$ & SSIM$\uparrow$ & LPIPS$\downarrow$\\\midrule
\multirow[c]{7}{*}[\dimexpr-\aboverulesep/2-\belowrulesep/2-\cmidrulewidth/2\relax]{\makebox[15pt][c]{\rotatebox[origin=c]{90}{\textbf{Phantom (P)}}}} & Cosmos-H & 56.72 & 65.58 & 54.12 & 67.21 & \underline{0.0234} & \underline{26.12} & \underline{0.8411} & \underline{0.0859} \\
 & BWM & \underline{57.89} & \underline{67.26} & \underline{56.77} & \underline{69.11} & 0.0266 & 25.83 & 0.8383 & \textbf{0.0796} \\
 & IRASim & 41.70 & 52.83 & 43.79 & 57.90 & 0.0383 & 23.58 & 0.7206 & 0.1566 \\
 & Ctrl-World & 51.51 & 61.73 & 49.60 & 62.93 & 0.0266 & 25.38 & 0.7974 & 0.1404 \\
 & SurgSora & 40.33 & 50.16 & 43.25 & 55.82 & 0.0353 & 23.48 & 0.7800 & 0.2283 \\
 & RoboStereo & 52.70 & 61.97 & 52.61 & 65.73 & 0.0248 & 25.64 & 0.8371 & 0.0919 \\
 & \textbf{SurgReact} & \textbf{60.31} & \textbf{69.17} & \textbf{57.39} & \textbf{70.27} & \textbf{0.0223} & \textbf{26.63} & \textbf{0.8466} & 0.0871 \\
\midrule
\multirow[c]{7}{*}[\dimexpr-\aboverulesep/2-\belowrulesep/2-\cmidrulewidth/2\relax]{\makebox[15pt][c]{\rotatebox[origin=c]{90}{\textbf{Ex vivo (E)}}}} & Cosmos-H & \underline{60.64} & \underline{70.15} & \underline{55.78} & 67.47 & \textbf{0.0208} & \textbf{26.83} & \textbf{0.8371} & \underline{0.0724} \\
 & BWM & 58.70 & 68.52 & 54.26 & 65.84 & 0.0238 & \underline{26.19} & 0.8294 & \textbf{0.0701} \\
 & IRASim & 52.59 & 64.40 & 47.96 & 62.56 & 0.0318 & 24.46 & 0.6940 & 0.1360 \\
 & Ctrl-World & 58.19 & 69.33 & 48.30 & 62.82 & 0.0256 & 25.50 & 0.7896 & 0.1471 \\
 & SurgSora & 47.67 & 60.05 & 43.21 & 58.06 & 0.0393 & 22.91 & 0.7551 & 0.2457 \\
 & RoboStereo & 58.89 & 69.94 & 55.66 & \underline{68.39} & 0.0232 & 24.74 & 0.7966 & 0.0814 \\
 & \textbf{SurgReact} & \textbf{60.95} & \textbf{72.60} & \textbf{58.19} & \textbf{71.49} & \underline{0.0226} & 25.81 & \underline{0.8335} & 0.0845 \\
\bottomrule\end{tabular}\end{adjustbox}
\par\medskip
\textbf{(b) Long window}\par\smallskip
\begin{adjustbox}{max width=\linewidth}\begin{tabular}{@{}cl*{8}{c}@{}}\toprule
& \multirow{2}{*}{\textbf{Method}} & \multicolumn{2}{c}{\textbf{Instr. Motion}} & \multicolumn{2}{c}{\textbf{Tissue Resp.}} & \multicolumn{4}{c}{\textbf{Visual Quality}}\\
\cmidrule(lr){3-4}\cmidrule(lr){5-6}\cmidrule(l){7-10}
&& AJ$\uparrow$ & $\delta_{\mathrm{avg}}^{\mathrm{vis}}\uparrow$ & AJ$\uparrow$ & $\delta_{\mathrm{avg}}^{\mathrm{vis}}\uparrow$ & L1$\downarrow$ & PSNR$\uparrow$ & SSIM$\uparrow$ & LPIPS$\downarrow$\\\midrule
\multirow[c]{7}{*}[\dimexpr-\aboverulesep/2-\belowrulesep/2-\cmidrulewidth/2\relax]{\makebox[15pt][c]{\rotatebox[origin=c]{90}{\textbf{Phantom (P)}}}} & Cosmos-H & 43.79 & 56.25 & 51.29 & 68.65 & \underline{0.0331} & \underline{23.68} & \underline{0.7834} & 0.1344 \\
 & BWM & \underline{44.24} & \underline{57.20} & \underline{53.11} & \underline{69.92} & 0.0375 & 23.15 & 0.7790 & \textbf{0.1214} \\
 & IRASim & 32.01 & 44.10 & 39.83 & 58.91 & 0.0565 & 20.93 & 0.6455 & 0.2113 \\
 & Ctrl-World & 40.21 & 53.48 & 45.77 & 65.18 & 0.0355 & 23.02 & 0.7403 & 0.1820 \\
 & SurgSora & 31.75 & 43.91 & 42.95 & 61.15 & 0.0585 & 20.46 & 0.6998 & 0.3186 \\
 & RoboStereo & 40.49 & 52.85 & 49.15 & 66.69 & 0.0354 & 23.18 & 0.7764 & 0.1505 \\
 & \textbf{SurgReact} & \textbf{50.80} & \textbf{63.15} & \textbf{56.23} & \textbf{72.41} & \textbf{0.0285} & \textbf{24.82} & \textbf{0.8069} & \underline{0.1216} \\
\midrule
\multirow[c]{7}{*}[\dimexpr-\aboverulesep/2-\belowrulesep/2-\cmidrulewidth/2\relax]{\makebox[15pt][c]{\rotatebox[origin=c]{90}{\textbf{Ex vivo (E)}}}} & Cosmos-H & \underline{50.68} & \underline{62.32} & 42.74 & 57.73 & \underline{0.0265} & \textbf{25.13} & \underline{0.8107} & \underline{0.1069} \\
 & BWM & 47.64 & 59.01 & 41.58 & 56.63 & 0.0294 & 24.31 & 0.8057 & \textbf{0.1023} \\
 & IRASim & 39.32 & 51.81 & 38.06 & 53.95 & 0.0432 & 22.61 & 0.6303 & 0.1656 \\
 & Ctrl-World & 46.43 & 60.53 & 39.74 & 57.39 & 0.0283 & 24.48 & 0.7878 & 0.1551 \\
 & SurgSora & 35.47 & 49.84 & 36.13 & 52.07 & 0.0877 & 18.23 & 0.6320 & 0.3052 \\
 & RoboStereo & 47.95 & 60.53 & \underline{44.19} & \underline{59.92} & 0.0296 & 23.23 & 0.7724 & 0.1217 \\
 & \textbf{SurgReact} & \textbf{51.57} & \textbf{64.93} & \textbf{47.76} & \textbf{64.19} & \textbf{0.0264} & \underline{24.62} & \textbf{0.8240} & 0.1082 \\
\bottomrule\end{tabular}\end{adjustbox}
\par\medskip
\end{table}

\subsection{Comparison with Existing Methods}

\noindent\textbf{Quantitative Comparison.}
Table~\ref{tab:interaction_comparison} compares instrument motion, tissue response, and visual quality across both domains and prediction windows. SurgReact achieves the highest AJ and $\delta_{\mathrm{avg}}^{\mathrm{vis}}$ for both instrument and tissue trajectories in all evaluated settings. In the phantom domain, it also achieves the best L1, PSNR, and SSIM under both prediction windows, showing that accurate interaction prediction is accompanied by strong visual quality. In the long-window setting, SurgReact obtains tissue AJ and visible position accuracy of 56.23 and 72.41, respectively, compared with 53.11 and 69.92 for BWM. On ex vivo data, Cosmos-H achieves better scores on all four reported visual metrics in the short-window setting and retains better PSNR and LPIPS in the long-window setting. Its extensive surgical-domain training, covering 32 datasets across nine robotic platforms~\cite{nelson2026open}, may contribute to its strong appearance generation. Nevertheless, SurgReact achieves the highest instrument and tissue trajectory scores in both ex vivo settings. In the long-window setting, its tissue AJ and visible position accuracy reach 47.76 and 64.19, respectively, compared with 42.74 and 57.73 for Cosmos-H. These results highlight SurgReact's strength in predicting tissue responses alongside instrument motion, providing interaction accuracy that is not fully reflected by appearance metrics alone.

\noindent\textbf{Qualitative Comparison.}
Figure~\ref{fig:interaction_comparison} presents five surgical scenarios of increasing interaction complexity, with zoomed-in regions highlighting instrument--tissue contact. In (a), baselines exhibit instrument shape distortions, misplaced instruments, or excessive tissue stretching, whereas SurgReact preserves both the grasping configuration and the corresponding tissue response. Scenarios (b--c) further distinguish visual realism from interaction accuracy: Cosmos-H and BWM generate realistic visuals in (b), but their instruments deviate from the reference positions. In (c), baselines depict unintended tissue lifting or disrupt local appearance, while our method preserves the instrument's proximity to the tissue surface.

Scenarios (d--e) place greater demands on coordinated instrument motion and tissue deformation. In (d), Cosmos-H and Ctrl-World largely reproduce the instrument positions but miss the corresponding soft-tissue response, showing that accurate positioning alone does not ensure correct interaction. Under the larger tissue deformation in (e), SurgReact maintains reference-consistent positions and more accurately reproduces the elongated tissue strip, whereas all baselines fail to capture this reaction coherently. This result highlights how conditioning tissue prediction on future instrument geometry helps maintain the spatial relationship between the grasping instruments and the stretched tissue, even under large deformation.

\begin{figure}[tbp]
\centering

\includegraphics[width=0.85\linewidth]{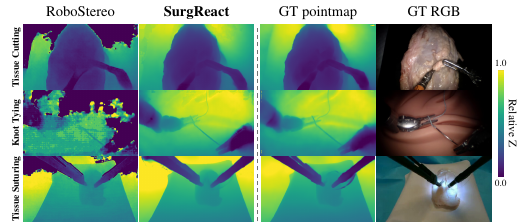}

\caption{{\textbf{Qualitative comparison of predicted pointmaps}, visualized using their relative depth ($Z$) component.}}
\label{fig:pointmap_comparison}

\end{figure}

Figure~\ref{fig:pointmap_comparison} further examines these differences in geometry. Our method preserves clear instrument contours, continuous tissue surfaces, and instrument--tissue boundaries in tissue cutting and suturing, while retaining slender structures and tissue folds in knot tying. In contrast, RoboStereo, which jointly predicts pointmaps and RGB, exhibits pronounced geometric noise, fragmented surfaces, and disrupted boundaries. This comparison highlights the value of explicitly modeling interaction geometry: SurgReact first predicts instrument geometry to guide tissue reaction prediction, providing faithful geometry for subsequent RGB generation.

\subsection{Ablation Studies}
\label{sec:ablations}

We conduct ablation studies to evaluate the main designs of SurgReact, we first introduce \emph{Direct}, which predicts future scene pointmaps from history geometry and actions, to evaluate our instrument-to-tissue prediction design. The variant w/o {ITP} removes Instrument Trajectory Prediction and initializes future predictions with current-frame instrument geometry and features, while w/o {CR} retains trajectory guidance but directly predicts future pointmaps instead of reference-based residuals.

As shown in Table~\ref{tab:ablation_geometry}, all variants degrade geometric accuracy and video quality. \emph{Direct} performs worst, suggesting that jointly learning instrument motion and tissue deformation from history geometry and actions is challenging. Its higher instrument and interaction errors support explicitly establishing future instrument geometry to condition tissue reaction prediction. Removing ITP also increases geometric errors, demonstrating the importance of anticipating future instrument placement rather than initializing predictions with current-frame geometry. Without CR, instrument trajectory guidance still improves performance over \emph{Direct}, indicating that future instrument geometry provides useful conditioning even when pointmaps are predicted directly. However, this variant remains less accurate than the full model, showing that reference-based residual prediction complements instrument guidance by learning coordinate corrections relative to the selected surface reference.

We evaluate \emph{Geometry and Feature Initialization} (GFI) by disabling it while retaining \emph{Reference Selection} (RS). This variant encodes history pointmaps and future pointmaps composed from instrument and tissue geometry using a shared encoder before joint spatiotemporal processing. We further disable RS by replacing learned selection with a nearest-depth rule. As shown in Table~\ref{tab:ablation_reference}, removing GFI increases coordinate and surface-normal errors and reduces video quality, supporting history encoding followed by motion-aligned feature initialization. Further disabling RS degrades performance, indicating that learned selection provides a better starting surface for coordinate refinement than fixed depth-based composition.

We evaluate teacher-student distillation by comparing Teacher, Student, and w/ Distillation. Teacher uses GT pointmaps as geometric conditions and serves as a reference. Student and w/ Distillation both use predicted pointmaps, with teacher supervision applied only to the latter.
As shown in Table~\ref{tab:ablation_distillation}, w/ Distillation improves all four visual metrics over Student, demonstrating that distillation helps SurgReact use imperfect predicted geometry more effectively. The remaining gap to Teacher highlights the importance of accurate geometric conditions and the potential for further gains from improved pointmap prediction.

\begin{table}[tbp]
\centering
\caption{{Ablation study} on \textbf{instrument guidance and coordinate refinement}.}
\label{tab:ablation_geometry}
\small
\setlength{\tabcolsep}{2pt}
\renewcommand{\arraystretch}{1.12}

\adjustbox{max width=\linewidth}{\begin{adjustbox}{max width=\linewidth}
\begin{tabular}{@{}*{12}{c}@{}}
\toprule
\multicolumn{2}{@{}l}{\multirow[c]{2}{*}[-2pt]{\textbf{Method}}}
& \multicolumn{2}{c}{\textbf{Global}}
& \multicolumn{2}{c}{\textbf{Instrument}}
& \multicolumn{2}{c}{\textbf{Interaction}}
& \multicolumn{4}{c}{\textbf{Visual quality}} \\
\cmidrule(l{1pt}r{1pt}){3-4}
\cmidrule(l{1pt}r{1pt}){5-6}
\cmidrule(l{1pt}r{1pt}){7-8}
\cmidrule(l{1pt}){9-12}
& &
$E_{\mathrm{3D}}\downarrow$ & $E_{\mathrm n}\downarrow$
& $E_{\mathrm{3D}}\downarrow$ & $E_{\mathrm n}\downarrow$
& $E_{\mathrm{3D}}\downarrow$ & $E_{\mathrm n}\downarrow$
& L1$\downarrow$ & PSNR$\uparrow$
& SSIM$\uparrow$ & LPIPS$\downarrow$ \\
\midrule
\multicolumn{2}{@{}l}{\emph{Direct}}
& 31.71 & 28.41
& 44.14 & 42.29
& 28.21 & 26.08
& 0.0338 & 22.22 & 0.7948 & 0.1455 \\
\multicolumn{2}{@{}l}{w/o ITP}
& 24.64 & 17.37
& 33.17 & 31.38
& 22.96 & 20.02
& 0.0308 & 23.04 & 0.8043 & 0.1330 \\
\multicolumn{2}{@{}l}{w/o CR}
& 30.62 & 27.84
& 39.63 & 40.22
& 26.40 & 25.08
& 0.0289 & 23.39 & 0.8164 & 0.1111 \\
\multicolumn{2}{@{}l}{\textbf{SurgReact}}
& \textbf{22.37} & \textbf{16.72}
& \textbf{27.84} & \textbf{28.87}
& \textbf{20.31} & \textbf{18.88}
& \textbf{0.0256} & \textbf{25.44}
& \textbf{0.8444} & \textbf{0.1025} \\
\bottomrule
\end{tabular}
\end{adjustbox}}

\end{table}

\begin{table}[tbp]
\centering
\caption{{Ablation study} on \textbf{initialization and reference selection}.}
\label{tab:ablation_reference}
\small
\setlength{\tabcolsep}{2pt}
\renewcommand{\arraystretch}{1.12}

\adjustbox{max width=\linewidth}{\begin{adjustbox}{max width=\linewidth}
\begin{tabular}{@{}*{12}{c}@{}}
\toprule
\multirow[c]{2}{*}[\dimexpr-\aboverulesep/2-\belowrulesep/2-\cmidrulewidth/2\relax]{\textbf{GFI}}
& \multirow[c]{2}{*}[\dimexpr-\aboverulesep/2-\belowrulesep/2-\cmidrulewidth/2\relax]{\textbf{RS}}
& \multicolumn{2}{c}{\textbf{Global}}
& \multicolumn{2}{c}{\textbf{Instrument}}
& \multicolumn{2}{c}{\textbf{Interaction}}
& \multicolumn{4}{c}{\textbf{Visual quality}} \\
\cmidrule(l{1pt}r{1pt}){3-4}
\cmidrule(l{1pt}r{1pt}){5-6}
\cmidrule(l{1pt}r{1pt}){7-8}
\cmidrule(l{1pt}){9-12}
& &
$E_{\mathrm{3D}}\downarrow$ & $E_{\mathrm n}\downarrow$
& $E_{\mathrm{3D}}\downarrow$ & $E_{\mathrm n}\downarrow$
& $E_{\mathrm{3D}}\downarrow$ & $E_{\mathrm n}\downarrow$
& L1$\downarrow$ & PSNR$\uparrow$
& SSIM$\uparrow$ & LPIPS$\downarrow$ \\
\midrule
$\times$ & $\times$
& 23.70 & 17.44
& 31.85 & 30.92
& 21.45 & 19.84
& 0.0265 & 24.12 & 0.8209 & 0.1064 \\
$\times$ & $\checkmark$
& 23.42 & 16.94
& 29.65 & 29.56
& 21.10 & 19.19
& 0.0264 & 24.76 & 0.8328 & 0.1042 \\
$\checkmark$ & $\checkmark$
& \textbf{22.37} & \textbf{16.72}
& \textbf{27.84} & \textbf{28.87}
& \textbf{20.31} & \textbf{18.88}
& \textbf{0.0256} & \textbf{25.44}
& \textbf{0.8444} & \textbf{0.1025} \\
\bottomrule
\end{tabular}
\end{adjustbox}}
\end{table}

\begin{figure}[tbp]
\centering
\begin{minipage}[c]{0.46\columnwidth}
\centering
\captionsetup{singlelinecheck=false,justification=raggedright}
\captionof{table}{Ablation study on \textbf{distillation strategy}.}
\label{tab:ablation_distillation}
\small
\setlength{\tabcolsep}{1.8pt}
\renewcommand{\arraystretch}{1.7}
\adjustbox{max width=\linewidth}{%
\begin{tabular}{@{}lcccc@{}}
\toprule
\textbf{Method} & \textbf{L1}$\downarrow$ & \textbf{PSNR}$\uparrow$ & \textbf{SSIM}$\uparrow$ & \textbf{LPIPS}$\downarrow$ \\
\midrule
Teacher & \textbf{0.0171} & \textbf{27.91} & \textbf{0.8962} & \textbf{0.0642} \\
\midrule
Student & 0.0250 & 24.50 & 0.8457 & 0.0979 \\
\textbf{w/ Distillation} & 0.0219 & 25.22 & 0.8626 & 0.0864 \\
\bottomrule
\end{tabular}}
\end{minipage}\hfill
\begin{minipage}[c]{0.48\columnwidth}
\centering
\includegraphics[trim=5bp 0 0 0,clip,width=\linewidth]{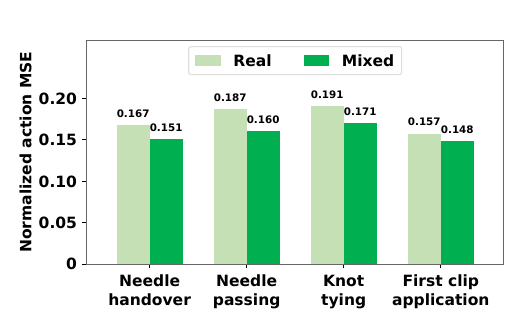}%
\par\vspace{1pt}
\captionsetup{singlelinecheck=false,justification=raggedright}
\caption{\textbf{Action prediction MSE.}}
\label{fig:policy_utility}
\end{minipage}
\end{figure}

\subsection{Discussion}
\label{sec:discussion}

\noindent\textbf{Policy Improvement with Generated Data.}
We test whether generated observations improve action prediction on real surgical videos. We use 300 real windows per task from Open-H-Embodiment, covering needle handover, needle passing, knot tying, and first clip application. SurgReact generates 300 corresponding windows conditioned on their visual histories and recorded actions, reusing these actions as training labels. We compare \textbf{Real} (real data only) with \textbf{Mixed} (a 1:1 real/generated mixture), using the same policy model GR00T-N1.6~\cite{bjorck2025gr00t} initialization and training budget.

As shown in Figure~\ref{fig:policy_utility}, Mixed reduces normalized action MSE on held-out real videos by 9.86\%, 14.40\%, 10.57\%, and 5.92\%, respectively. These gains support using SurgReact-generated observations to supplement real demonstrations for surgical action prediction.

\Needspace{12\baselineskip}
\section{Conclusion}
\label{sec:conclusion}

We presented SurgReact, a surgical world model that explicitly predicts instrument-conditioned tissue geometry before video synthesis. SurgReact first forecasts future instrument geometry, then combines it with tissue history to predict tissue reactions through reference selection and coordinate refinement. The resulting pointmaps guide video generation, while teacher-student distillation helps the generator effectively use predicted geometry. Experiments on phantom and ex vivo data demonstrate improved geometry and instrument-tissue interaction prediction together with visual quality. Additionally, SurgReact generated observations also improve downstream surgical action prediction. 
\FloatBarrier
{\fontsize{8.5}{10}\selectfont\linespread{1.0}\selectfont\setlength{\parskip}{0pt}\setlength{\bibsep}{0pt plus.3pt}
\bibliographystyle{ieeenat_fullname}
\bibliography{references}}

\begin{thebibliography}{40}
\providecommand{\natexlab}[1]{#1}
\providecommand{\url}[1]{\texttt{#1}}
\expandafter\ifx\csname urlstyle\endcsname\relax
  \providecommand{\doi}[1]{doi: #1}\else
  \providecommand{\doi}{doi: \begingroup \urlstyle{rm}\Url}\fi

\bibitem[Bjorck et~al.(2025)Bjorck, Casta{\~n}eda, Cherniadev, Da, Ding, Fan,
  Fang, Fox, Hu, Huang, et~al.]{bjorck2025gr00t}
Johan Bjorck, Fernando Casta{\~n}eda, Nikita Cherniadev, Xingye Da, Runyu Ding,
  Linxi Fan, Yu Fang, Dieter Fox, Fengyuan Hu, Spencer Huang, et~al.
\newblock {GR00T N1}: An open foundation model for generalist humanoid robots.
\newblock \emph{arXiv preprint arXiv:2503.14734}, 2025.

\bibitem[{BWM Team}(2026)]{bwm2026bwm}
{BWM Team}.
\newblock {BWM}: A low-cost high-fidelity world simulator for robot learning.
\newblock \emph{arXiv preprint arXiv:2607.29302}, 2026.

\bibitem[Cao et~al.(2025)Cao, Zhou, Li, Liang, Yu, Wang, Xue, and
  Fu]{cao2025uni3c}
Chenjie Cao, Jingkai Zhou, Shikai Li, Jingyun Liang, Chaohui Yu, Fan Wang,
  Xiangyang Xue, and Yanwei Fu.
\newblock {Uni3C}: Unifying precisely {3D}-enhanced camera and human motion
  controls for video generation.
\newblock In \emph{SIGGRAPH ASIA}, pages 1--12, 2025.

\bibitem[Carion et~al.(2026)Carion, Gustafson, Hu, Debnath, Hu, Suris
  Coll-Vinent, Ryali, Alwala, Khedr, Huang, Lei, Ma, Guo, Kalla, Marks, Greer,
  Wang, Sun, R\"{a}dle, Afouras, Mavroudi, Xu, Wu, Zhou, Momeni, Hazra, Ding,
  Vaze, Porcher, Li, Li, Kamath, Cheng, Dollar, Ravi, Saenko, Zhang, and
  Feichtenhofer]{ICLR2026_e0982cbc}
Nicolas Carion, Laura Gustafson, Yuan-Ting Hu, Shoubhik Debnath, Ronghang Hu,
  Didac Suris Coll-Vinent, Chaitanya Ryali, Kalyan~Vasudev Alwala, Haitham
  Khedr, Andrew Huang, Jie Lei, Tengyu Ma, Baishan Guo, Arpit Kalla, Markus
  Marks, Joseph Greer, Meng Wang, Peize Sun, Roman R\"{a}dle, Triantafyllos
  Afouras, Effrosyni Mavroudi, Katherine Xu, Tsung-Han Wu, Yu Zhou, Liliane
  Momeni, Rishi Hazra, Shuangrui Ding, Sagar Vaze, Francois Porcher, Feng Li,
  Siyuan Li, Aishwarya Kamath, Ho~Kei Cheng, Piotr Dollar, Nikhila Ravi, Kate
  Saenko, Pengchuan Zhang, and Christoph Feichtenhofer.
\newblock {SAM 3}: Segment anything with concepts.
\newblock In \emph{ICLR}, pages 138846--138923, 2026.

\bibitem[Chen et~al.(2025)Chen, Yang, Wang, Bai, Ren, and
  Zhou]{chen2025surgsora}
Tong Chen, Shuya Yang, Junyi Wang, Long Bai, Hongliang Ren, and Luping Zhou.
\newblock {SurgSora}: Object-aware diffusion model for controllable surgical
  video generation.
\newblock In \emph{MICCAI}, pages 521--531, 2025.

\bibitem[Chen et~al.(2018)Chen, Badrinarayanan, Drozdov, and
  Rabinovich]{chen2018estimating}
Zhao Chen, Vijay Badrinarayanan, Gilad Drozdov, and Andrew Rabinovich.
\newblock Estimating depth from {RGB} and sparse sensing.
\newblock In \emph{ECCV}, pages 176--192, 2018.

\bibitem[Dantas et~al.(2025)Dantas, Jorge, Nicolau, Vales, Coutinho, Rodrigues,
  Febra, and Lopes]{dantas2025video}
I. Dantas, I. Jorge, A. Nicolau, M. Vales, C. Coutinho, S. Rodrigues, P. Febra,
  and V.~N. Lopes.
\newblock Video-based learning for basic surgical skills - a randomized trial.
\newblock \emph{Surgery Open Science}, pages 94--106, 2025.

\bibitem[Doersch et~al.(2022)Doersch, Gupta, Markeeva, Recasens, Smaira, Aytar,
  Carreira, Zisserman, and Yang]{NEURIPS2022_58168e8a}
Carl Doersch, Ankush Gupta, Larisa Markeeva, Adria Recasens, Lucas Smaira,
  Yusuf Aytar, Joao Carreira, Andrew Zisserman, and Yi Yang.
\newblock {TAP-Vid}: A benchmark for tracking any point in a video.
\newblock In \emph{NeurIPS}, pages 13610--13626, 2022.

\bibitem[Guo et~al.(2026)Guo, Shi, Chen, and Finn]{guo2026ctrl}
Yanjiang Guo, Lucy Shi, Jianyu Chen, and Chelsea Finn.
\newblock {Ctrl-World}: A controllable generative world model for robot
  manipulation.
\newblock In \emph{ICLR}, pages 6121--6138, 2026.

\bibitem[Hu et~al.(2022)Hu, Shen, Wallis, Allen-Zhu, Li, Wang, Wang, and
  Chen]{hu2022lora}
Edward~J Hu, Yelong Shen, Phillip Wallis, Zeyuan Allen-Zhu, Yuanzhi Li, Shean
  Wang, Lu Wang, and Weizhu Chen.
\newblock Lo{RA}: Low-rank adaptation of large language models.
\newblock In \emph{ICLR}, 2022.

\bibitem[Huang et~al.(2025)Huang, Chen, Zhang, Sun, and
  Schwager]{pmlr-v305-huang25c}
Suning Huang, Qianzhong Chen, Xiaohan Zhang, Jiankai Sun, and Mac Schwager.
\newblock {ParticleFormer}: A {3D} point cloud world model for multi-object,
  multi-material robotic manipulation.
\newblock In \emph{CoRL}, pages 4941--4957, 2025.

\bibitem[Huang et~al.(2026)Huang, Chao, Mousavian, Liu, Fox, Mo, and
  Fei-Fei]{Huang_2026_CVPR}
Wenlong Huang, Yu-Wei Chao, Arsalan Mousavian, Ming-Yu Liu, Dieter Fox, Kaichun
  Mo, and Li Fei-Fei.
\newblock {PointWorld}: Scaling {3D} world models for in-the-wild robotic
  manipulation.
\newblock In \emph{CVPR}, pages 20765--20779, 2026.

\bibitem[Jang et~al.(2025)Jang, Ye, Lin, Xiang, Bjorck, Fang, Hu, Huang,
  Kundalia, Lin, Magne, Mandlekar, Narayan, Tan, Wang, Wang, Wang, Xu, Zeng,
  Zheng, Zheng, Liu, Zettlemoyer, Fox, Kautz, Reed, Zhu, and
  Fan]{pmlr-v305-jang25a}
Joel Jang, Seonghyeon Ye, Zongyu Lin, Jiannan Xiang, Johan Bjorck, Yu Fang,
  Fengyuan Hu, Spencer Huang, Kaushil Kundalia, Yen-Chen Lin, Lo\"{i}c Magne,
  Ajay Mandlekar, Avnish Narayan, You~Liang Tan, Guanzhi Wang, Jing Wang, Qi
  Wang, Yinzhen Xu, Xiaohui Zeng, Kaiyuan Zheng, Ruijie Zheng, Ming-Yu Liu,
  Luke Zettlemoyer, Dieter Fox, Jan Kautz, Scott Reed, Yuke Zhu, and Linxi Fan.
\newblock {DreamGen}: Unlocking generalization in robot learning through video
  world models.
\newblock In \emph{CoRL}, pages 5170--5194, 2025.

\bibitem[Karaev et~al.(2025)Karaev, Makarov, Wang, Neverova, Vedaldi, and
  Rupprecht]{Karaev_2025_ICCV}
Nikita Karaev, Yuri Makarov, Jianyuan Wang, Natalia Neverova, Andrea Vedaldi,
  and Christian Rupprecht.
\newblock {CoTracker3}: Simpler and better point tracking by pseudo-labelling
  real videos.
\newblock In \emph{ICCV}, pages 6013--6022, 2025.

\bibitem[Kim et~al.(2025)Kim, Chen, Hansen, Shi, Goldenberg, Schmidgall,
  Scheikl, Deguet, White, Tsai, Cha, Jopling, Finn, and
  Krieger]{doi:10.1126/scirobotics.adt5254}
Ji~Woong~(Brian) Kim, Juo-Tung Chen, Pascal Hansen, Lucy~Xiaoyang Shi, Antony
  Goldenberg, Samuel Schmidgall, Paul~Maria Scheikl, Anton Deguet, Brandon~M.
  White, De~Ru Tsai, Richard~Jaepyeong Cha, Jeffrey Jopling, Chelsea Finn, and
  Axel Krieger.
\newblock {SRT-H}: A hierarchical framework for autonomous surgery via
  language-conditioned imitation learning.
\newblock \emph{Science Robotics}, page eadt5254, 2025.

\bibitem[Li et~al.(2026)Li, Yang, Peng, Guo, Zhang, Tao, Duan, Xu, Dou, Jin,
  et~al.]{li2026articulated}
Bohan Li, Shuojue Yang, Baorui Peng, Xianda Guo, Erli Zhang, Youqi Tao, Junfeng
  Duan, Daguang Xu, Qi Dou, Xin Jin, et~al.
\newblock From articulated kinematics to routed visual control for
  action-conditioned surgical video generation.
\newblock In \emph{NeurIPS}, 2026.

\bibitem[Li et~al.(2019)Li, Wu, Tedrake, Tenenbaum, and
  Torralba]{li2019learning}
Yunzhu Li, Jiajun Wu, Russ Tedrake, Joshua~B. Tenenbaum, and Antonio Torralba.
\newblock Learning particle dynamics for manipulating rigid bodies, deformable
  objects, and fluids.
\newblock In \emph{ICLR}, 2019.

\bibitem[Lin et~al.(2026)Lin, Chen, Liew, Chen, Li, Zhao, Peng, Guo, Zhou, Shi,
  Feng, and Kang]{ICLR2026_e4cd5012}
Haotong Lin, Sili Chen, Jun~Hao Liew, Donny~Y. Chen, Zhenyu Li, Yang Zhao, Sida
  Peng, Hengkai Guo, Xiaowei Zhou, Guang Shi, Jiashi Feng, and Bingyi Kang.
\newblock {Depth Anything 3}: Recovering the visual space from any views.
\newblock In \emph{ICLR}, pages 141261--141285, 2026.

\bibitem[Lipman et~al.(2023)Lipman, Chen, Ben-Hamu, Nickel, and
  Le]{lipman2023flow}
Yaron Lipman, Ricky T.~Q. Chen, Heli Ben-Hamu, Maximilian Nickel, and Matthew
  Le.
\newblock Flow matching for generative modeling.
\newblock In \emph{ICLR}, 2023.

\bibitem[Liu et~al.(2025)Liu, Wang, Mishra, Gao, Qin, Low, Kong, Zhuo, Fu, Ng,
  and Jin]{liu2025surgslot}
Haofeng Liu, Ziyue Wang, Sudhanshu Mishra, Mingqi Gao, Guanyi Qin, Chang~Han
  Low, Alex Y.~W. Kong, Zhu Zhuo, Huazhu Fu, Joseph~S. Ng, and Yueming Jin.
\newblock {SurgSLOT}: Segment anything in surgical videos via semantic
  long-term tracking.
\newblock \emph{arXiv preprint arXiv:2511.16618}, 2025.

\bibitem[Liu et~al.(2026)Liu, Zhou, Shao, Lin, Li, Lin, Ma, Yang, Tang, Chen,
  et~al.]{liu2026surgcast}
Wanhao Liu, Rulin Zhou, Liangjing Shao, Zhaocheng Lin, Dongyue Li, Jinsong Lin,
  Guoheng Ma, Hangheng Yang, Zhiqing Tang, Jing Chen, et~al.
\newblock {SurgCast}: Action-conditioned future skeletons for controllable
  surgical world models.
\newblock \emph{arXiv preprint arXiv:2610.15645}, 2026.

\bibitem[Lopez-Paz et~al.(2016)Lopez-Paz, Bottou, Sch{\"o}lkopf, and
  Vapnik]{lopezpaz2016unifying}
David Lopez-Paz, L{\'e}on Bottou, Bernhard Sch{\"o}lkopf, and Vladimir Vapnik.
\newblock Unifying distillation and privileged information.
\newblock In \emph{ICLR}, 2016.

\bibitem[Nelson et~al.(2026)Nelson, Chen, Haworth, Chen, Zbinden, Huang,
  Abdelaal, Arezzo, Acar, Alambeigi, et~al.]{nelson2026open}
Nigel Nelson, Juo-Tung Chen, Jesse Haworth, Xinhao Chen, Lukas Zbinden, Dianye
  Huang, Alaa~Eldin Abdelaal, Alberto Arezzo, Ayberk Acar, Farshid Alambeigi,
  et~al.
\newblock {Open-H-Embodiment}: A large-scale dataset for enabling foundation
  models in medical robotics.
\newblock \emph{arXiv preprint arXiv:2604.21017}, 2026.

\bibitem[Nwoye et~al.(2022)Nwoye, Yu, Gonzalez, Seeliger, Mascagni, Mutter,
  Marescaux, and Padoy]{nwoye2022rendezvous}
Chinedu~Innocent Nwoye, Tong Yu, Cristians Gonzalez, Barbara Seeliger, Pietro
  Mascagni, Didier Mutter, Jacques Marescaux, and Nicolas Padoy.
\newblock {Rendezvous}: Attention mechanisms for the recognition of surgical
  action triplets in endoscopic videos.
\newblock \emph{Medical Image Analysis}, page 102433, 2022.

\bibitem[Pan et~al.(2026)Pan, Li, Liu, Liu, Liu, and Yuan]{pan2026surgvista}
Wentao Pan, Wuyang Li, Shengyuan Liu, Xinyu Liu, Hengyu Liu, and Yixuan Yuan.
\newblock {SurgVista}: Long-horizon surgical world modeling with plausible
  instrument-tissue dynamics.
\newblock In \emph{NeurIPS}, 2026.

\bibitem[Peri et~al.(2026)Peri, Nguyen, Kim, Fuxin, and Lee]{peri20263d}
Skand Peri, Hung Nguyen, Chanho Kim, Li Fuxin, and Stefan Lee.
\newblock {3D} point world models: Point completion enables more accurate
  dynamics learning.
\newblock \emph{arXiv preprint arXiv:2607.00148}, 2026.

\bibitem[Purica et~al.(2015)Purica, Mora, Pesquet-Popescu, Cagnazzo, and
  Ionescu]{purica2015improved}
Andrei~I Purica, Elie~G Mora, Beatrice Pesquet-Popescu, Marco Cagnazzo, and
  Bogdan Ionescu.
\newblock Improved view synthesis by motion warping and temporal hole filling.
\newblock In \emph{ICASSP}, pages 1191--1195, 2015.

\bibitem[Romero et~al.(2015)Romero, Ballas, Ebrahimi~Kahou, Chassang, Gatta,
  and Bengio]{romero2015fitnets}
Adriana Romero, Nicolas Ballas, Samira Ebrahimi~Kahou, Antoine Chassang, Carlo
  Gatta, and Yoshua Bengio.
\newblock {FitNets}: Hints for thin deep nets.
\newblock In \emph{ICLR}, 2015.

\bibitem[Shi et~al.(2022)Shi, Xu, Huang, Li, and Wu]{shi2022robocraft}
Haochen Shi, Huazhe Xu, Zhiao Huang, Yunzhu Li, and Jiajun Wu.
\newblock {RoboCraft}: Learning to see, simulate, and shape elasto-plastic
  objects with graph networks.
\newblock In \emph{Robotics: Science and Systems}, 2022.

\bibitem[Twinanda et~al.(2017)Twinanda, Shehata, Mutter, Marescaux,
  de~Mathelin, and Padoy]{7519080}
Andru~P. Twinanda, Sherif Shehata, Didier Mutter, Jacques Marescaux, Michel de
  Mathelin, and Nicolas Padoy.
\newblock {EndoNet}: A deep architecture for recognition tasks on laparoscopic
  videos.
\newblock \emph{IEEE TMI}, pages 86--97, 2017.

\bibitem[Wan et~al.(2025)Wan, Wang, Ai, Wen, Mao, Xie, Chen, Yu, Zhao, Yang,
  et~al.]{wan2025wan}
Team Wan, Ang Wang, Baole Ai, Bin Wen, Chaojie Mao, Chen-Wei Xie, Di Chen,
  Feiwu Yu, Haiming Zhao, Jianxiao Yang, et~al.
\newblock {Wan}: Open and advanced large-scale video generative models.
\newblock \emph{arXiv preprint arXiv:2503.20314}, 2025.

\bibitem[Wang et~al.(2024)Wang, Leroy, Cabon, Chidlovskii, and
  Revaud]{wang2024dust3r}
Shuzhe Wang, Vincent Leroy, Yohann Cabon, Boris Chidlovskii, and Jerome Revaud.
\newblock {DUSt3R}: Geometric {3D} vision made easy.
\newblock In \emph{CVPR}, pages 20697--20709, 2024.

\bibitem[Wang et~al.(2004)Wang, Bovik, Sheikh, and Simoncelli]{1284395}
Zhou Wang, A.C. Bovik, H.R. Sheikh, and E.P. Simoncelli.
\newblock Image quality assessment: from error visibility to structural
  similarity.
\newblock \emph{IEEE TIP}, pages 600--612, 2004.

\bibitem[Xu et~al.(2026{\natexlab{a}})Xu, Zhang, Liu, Chen, Han, and
  Liu]{Xu_2026_CVPR}
Mutian Xu, Tianbao Zhang, Tianqi Liu, Zhaoxi Chen, Xiaoguang Han, and Ziwei
  Liu.
\newblock {Kinema4D}: Kinematic {4D} world modeling for spatiotemporal embodied
  simulation.
\newblock In \emph{CVPRW}, pages 4495--4500, 2026{\natexlab{a}}.

\bibitem[Xu et~al.(2026{\natexlab{b}})Xu, Liang, Han, Xi, Rao, Zhou, Zhang,
  Yan, Wei, Huang, et~al.]{xu2026causalwm}
Ziming Xu, Shuang Liang, Ruobing Han, Ziqiao Xi, Mingxing Rao, Kun Zhou, Zijun
  Zhang, Yuchen Yan, Yufan Wei, Junbo Huang, et~al.
\newblock {CausalWM}: Causal chain-of-thought reasoning for embodied world
  model.
\newblock \emph{arXiv preprint arXiv:2609.23184}, 2026{\natexlab{b}}.

\bibitem[Yang et~al.(2024)Yang, Du, Ghasemipour, Tompson, Kaelbling,
  Schuurmans, and Abbeel]{ICLR2024_c4d66eae}
Sherry Yang, Yilun Du, Seyed Ghasemipour, Jonathan Tompson, Leslie Kaelbling,
  Dale Schuurmans, and Pieter Abbeel.
\newblock Learning interactive real-world simulators.
\newblock In \emph{ICLR}, pages 45210--45234, 2024.

\bibitem[Zhang et~al.(2023)Zhang, Rao, and Agrawala]{zhang2023controlnet}
Lvmin Zhang, Anyi Rao, and Maneesh Agrawala.
\newblock Adding conditional control to text-to-image diffusion models.
\newblock In \emph{ICCV}, pages 3836--3847, 2023.

\bibitem[Zhang et~al.(2018)Zhang, Isola, Efros, Shechtman, and
  Wang]{zhang2018unreasonable}
Richard Zhang, Phillip Isola, Alexei~A Efros, Eli Shechtman, and Oliver Wang.
\newblock The unreasonable effectiveness of deep features as a perceptual
  metric.
\newblock In \emph{CVPR}, pages 586--595, 2018.

\bibitem[Zhang et~al.(2026)Zhang, Chen, Xu, Liu, Zhong, Zhang, Zhou, and
  Li]{zhang2026robostereo}
Ruicheng Zhang, Guangyu Chen, Zunnan Xu, Zihao Liu, Zhizhou Zhong, Mingyang
  Zhang, Jun Zhou, and Xiu Li.
\newblock {RoboStereo}: Dual-tower {4D} embodied world models for unified
  policy optimization.
\newblock \emph{arXiv preprint arXiv:2603.12639}, 2026.

\bibitem[Zhu et~al.(2025)Zhu, Wu, Guo, Liu, Cheang, and Kong]{zhu2025irasim}
Fangqi Zhu, Hongtao Wu, Song Guo, Yuxiao Liu, Chilam Cheang, and Tao Kong.
\newblock {IRASim}: A fine-grained world model for robot manipulation.
\newblock In \emph{ICCV}, pages 9834--9844, 2025.

\end{thebibliography}
\clearpage\appendix
\linespread{.97}\selectfont
\setlength{\parskip}{2pt plus.5pt minus.5pt}
\setlength{\textfloatsep}{9pt plus2pt minus1pt}
\setlength{\floatsep}{8pt plus2pt minus1pt}
\setlength{\intextsep}{8pt plus2pt minus1pt}
\setcounter{figure}{0}\setcounter{table}{0}\setcounter{equation}{0}
\renewcommand{\thefigure}{S\arabic{figure}}\renewcommand{\thetable}{S\arabic{table}}\renewcommand{\theequation}{S\arabic{equation}}
\renewcommand{\theHfigure}{supp.\arabic{figure}}\renewcommand{\theHtable}{supp.\arabic{table}}\renewcommand{\theHequation}{supp.\arabic{equation}}
{\titlefont Supplementary Material\par}\medskip

\setcounter{section}{0}
\setcounter{figure}{0}
\setcounter{table}{0}
\setcounter{equation}{0}

\noindent\textbf{Overview.}
References without an ``S'' prefix refer to the main paper, such as Sec.~3.2, Eq.~(3), Table~1, and Fig.~4, while prefixes \texttt{A--E} and \texttt{S1--S8} refer to this supplement:
\begin{itemize}[leftmargin=1.1em, itemsep=1.5pt, topsep=2.5pt]
    \item \textbf{Sec.~\ref{sec:supp_method}:} Input parameterizations, progressive training schedules, loss formulations, and hyperparameters (Table~\ref{tab:supp_hyperparams}).
    \item \textbf{Sec.~\ref{sec:supp_setup}:} Dataset splits (Table~\ref{tab:supp_datasets}), baseline setups (Table~\ref{tab:supp_baselines}), evaluation metrics, and controlled ablation and policy protocols (Table~\ref{tab:supp_ablation_matrix}).
    \item \textbf{Sec.~\ref{sec:supp_rollout}:} Long-horizon 21.33-s autoregressive rollout protocol, quantitative benchmark (Table~\ref{tab:supp_rollout}), and qualitative rollouts (Figs.~\ref{fig:supp_rollout_w24}--\ref{fig:supp_rollout_w28}).
    \item \textbf{Sec.~\ref{sec:supp_diagnostics}:} Pointmap accuracy against RoboStereo (Table~\ref{tab:supp_robostereo_geo}), stage-wise geometric refinement and inference latency (Tables~\ref{tab:supp_coarse_to_fine}--\ref{tab:supp_runtime}, Fig.~\ref{fig:supp_coarse_to_fine}), qualitative component ablations for Table~2 (Fig.~\ref{fig:supp_ablation_t2}), and video comparisons on the \href{https://GenerativeIntelligence-HITSZ.github.io/SurgReact/}{project webpage}.
    \item \textbf{Sec.~\ref{sec:supp_limitations}:} Limitations and future work.
\end{itemize}

\section{Method Details}
\label{sec:supp_method}

\subsection{Input Parameterizations (Secs.~3.1--3.4)}
\label{sec:supp_inputs}

\noindent Throughout this supplement, $N{=}5$ history frames are indexed by $t{\in}\{-4,\dots,0\}$, with $t{=}0$ denoting the current frame, and future frames by $t{\in}\{1,\dots,T\}$ for $T{\in}\{28,80\}$ at 15~Hz, i.e., 15~fps.

\modhead{Action Representation}
At each timestep, the action vector $a_t$ encodes dual-arm kinematics from Open-H-Embodiment~\cite{nelson2026open}. Each arm contributes 3D translation relative to $t{=}0$ in the source robot frame, a 6D relative rotation given by the first two columns of the relative rotation matrix, and 1D absolute jaw opening, totaling 20 kinematic dimensions. These continuous channels are concatenated with relative timing cues and per-channel binary validity flags to distinguish missing recordings from physical zeros across heterogeneous platforms. Valid continuous channels are standardized per domain and channel using training-set statistics. Stereo views share identical actions, and no robot--camera hand--eye calibration is required.

\modhead{Pointmap Alignment and Normalization}
History and reference pointmaps, together with per-frame camera intrinsics and world-to-camera extrinsics $\mathbf{W}_t$, are estimated directly from RGB frames using Depth Anything~3 (DA3)~\cite{ICLR2026_e4cd5012} on native half-resolution grids, preserving original aspect ratios and camera rays without external camera calibration. Crucially, for all end-to-end inference and benchmark evaluations (Table~1, Table~\ref{tab:supp_rollout}, and Table~\ref{tab:supp_runtime}) across both short and long horizons, conditioning history pointmaps $P_{\mathrm{his}}$ are reconstructed online in a history-only DA3 pass using strictly the five observed RGB frames ($t{\in}\{-4,\dots,0\}$), ensuring that all history coordinates, camera transforms, and per-window normalization statistics depend solely on observed frames with zero future-frame leakage. During offline training and controlled geometric ablations (Tables~2--4 and Tables~\ref{tab:supp_robostereo_geo}--\ref{tab:supp_coarse_to_fine}), conditioning history and reference pointmaps ($P_{\mathrm{his}}, P_{1:T}^*$) for all evaluated variants are drawn from a shared full-window DA3 reconstruction anchored at $t{=}0$, ensuring an identical camera gauge without inter-pass monocular scale drift; future reference pointmaps are never accessed by SurgReact at inference (except in the reference-geometry oracle diagnostics: \emph{Teacher} in Table~4 and \emph{SurgReact (Ref.\ Geo.)} in Table~\ref{tab:supp_rollout}). Within each DA3 reconstruction, all 3D points are transformed into the physical $t{=}0$ camera frame $C_0$ via $X_t^{C_0}{=}\mathbf{W}_0 \mathbf{W}_t^{-1} X_t^{C_t}$ and normalized as $P_t{=}(X_t^{C_0}-c_P)/s_P$, where $c_P$ is the per-axis median of valid coordinates at $t{=}0$ and $s_P$ is their median depth. Before encoding with the frozen video VAE, pointmaps are scaled to $[-1,1]$ using per-domain 0.5th--99.5th training percentiles and letterboxed to $512{\times}288$ to match the RGB input.

\modhead{Instrument Masks, Keypoints, and Tissue Reference}
History instrument masks $M_{\mathrm{his}}$ are extracted with SAM~3.1~\cite{ICLR2026_e0982cbc} and refined across the five observed frames with SurgSLOT~\cite{liu2025surgslot}. For each visible instrument $k$, the anchor keypoint $u_0^k$ is placed at the skeleton midpoint of $M_0^k$, or at an interior pixel when the midpoint falls outside the mask, and tracked over the history window with CoTracker3~\cite{Karaev_2025_ICCV}. When transported instruments overlap in Eq.~(3), the smaller valid depth is retained, while inactive slots are masked using separate 2D and 3D validity flags. To construct the tissue reference $B$ and aligned features $E^{\mathrm{tissue}}$ in Eq.~(5), valid non-instrument history points are reprojected onto the current image grid, prioritizing the most recent frame from $t{=}0$ back to $t{=}-4$ and the nearest positive depth within each frame without depth averaging. Remaining occluded pixels are filled by backprojecting the nearest valid neighboring tissue depth along each pixel's camera ray.

\subsection{Progressive Training Schedules (Secs.~3.2--3.4)}
\label{sec:supp_progressive}

\modhead{Stage 1: Geometry Predictor}
In Eqs.~(7), (8), and (10), binary availability and source-status channels indicate whether each pixel originates from valid instrument geometry, directly observed tissue, or neighbor-filled tissue; these channels are concatenated with geometry features to guide spatiotemporal reasoning $\mathcal{T}$, reference selection $\mathcal{C}_{\mathrm{sel}}$, and coordinate refinement $\mathcal{F}_{\mathrm{refine}}$. The geometry predictor of Secs.~3.2--3.3 is trained in three phases:
\begin{enumerate}
    \item \emph{Trajectory warmup:} pretrain the instrument trajectory predictor alone using $\mathcal{L}_{\mathrm{pos}}$.
    \item \emph{Tissue-reaction warmup:} freeze the trajectory predictor and train the scene encoder, spatiotemporal Transformer, selector, and refinement network under stable surface transport.
    \item \emph{End-to-end Stage-1 fine-tuning:} jointly optimize all Stage-1 modules under $\mathcal{L}_{\mathrm{geo}}$ with a geometry learning rate of $10^{-4}$ and a trajectory learning rate of $10^{-5}$.
\end{enumerate}
Stage~1 is decoupled from Stage~2, so video generation losses are never backpropagated into the geometry predictor.

\modhead{Stage 2: Video Generator and Distillation}
Built on Wan2.2 TI2V-5B~\cite{wan2025wan} at $512{\times}288$ with an empty text prompt, the 8-layer control Transformer $\mathcal{C}_\phi$ injects zero-initialized geometry residuals into $J{=}8$ evenly spaced backbone blocks. No kinematic actions are fed directly to the video generator, and both the base Wan2.2 weights and video VAE remain frozen throughout training. Adaptation proceeds in two phases:
\begin{enumerate}
    \item \emph{Control warmup:} train $\mathcal{C}_\phi$ alone while keeping backbone LoRA frozen so that early control residuals do not perturb the pretrained video prior.
    \item \emph{Joint control and LoRA:} jointly optimize $\mathcal{C}_\phi$ at a reduced learning rate together with backbone LoRA~\cite{hu2022lora} of rank $r{=}64$ and $\alpha{=}32$ on attention and FFN projections.
\end{enumerate}
During distillation in Sec.~3.4, the frozen reference-pointmap teacher supervises the predicted-pointmap student, and only the student is used at inference.

\subsection{Loss Formulations and Hyperparameters (Secs.~3.2--3.4)}
\label{sec:supp_losses}

\modhead{Stage-1 Geometry Losses}
In Eq.~(12), the trajectory loss $\mathcal{L}_{\mathrm{pos}}$ averages per-coordinate $\ell_1$ errors over valid future 2D pseudo-labels from CoTracker3 and 3D pseudo-labels from DA3, where $\operatorname{Mean}_{\mathrm{uv}}$ and $\operatorname{Mean}_{\mathrm{xyz}}$ average over valid future $(t,k)$ entries:
\begin{align}
\nonumber
\mathcal{L}_{\mathrm{pos}}
&= \operatorname{Mean}_{\mathrm{uv}}\!\left(\tfrac{1}{2}\|\widehat{u}_t^k - u_t^{k,*}\|_1\right) \\
\label{eq:supp_lpos}
&\quad + \operatorname{Mean}_{\mathrm{xyz}}\!\left(\tfrac{1}{3}\|\widehat{x}_t^k - x_t^{k,*}\|_1\right).
\end{align}
The region-weighted coordinate loss $\mathcal{L}_{\mathrm{xyz}}$ and multiscale surface shape loss $\mathcal{L}_{\mathrm{surf}}$ over horizontal and vertical pixel separations $\mathcal{R}{=}\{1,2,4,8\}$ are defined as
\begin{align}
\label{eq:supp_lxyz}
\mathcal{L}_{\mathrm{xyz}}
&= \frac{\sum_{(t,u)\in\Omega} w_t(u)\,\|\widehat{P}_t(u) - P_t^*(u)\|_1}{3\sum_{(t,u)\in\Omega} w_t(u)}, \\
\label{eq:supp_lsurf}
\mathcal{L}_{\mathrm{surf}}
&= \frac{1}{|\mathcal{R}|}\sum_{d\in\mathcal{R}} \mathcal{L}_{\mathrm{surf}}^{(d)}, \\
\nonumber
\mathcal{L}_{\mathrm{surf}}^{(d)}
&= \frac{\sum_{(t,u,v)\in\Omega_d} \bar{w}_t(u,v)\,
\|\delta\widehat{P}_t(u,v) - \delta P_t^*(u,v)\|_1}{3\sum_{(t,u,v)\in\Omega_d}\bar{w}_t(u,v)},
\end{align}
where $\Omega$ denotes valid future pixel locations $(t,u)$ with finite $P_t^*(u)$, $\Omega_d$ denotes valid horizontal and vertical pixel pairs $(t,u,v)$ separated by $d$ pixels with both endpoints in $\Omega$, $\delta P_t(u,v) = P_t(v) - P_t(u)$, $\bar{w}_t(u,v) = \tfrac{1}{2}[w_t(u) + w_t(v)]$, and $w_t(u)$ assigns detached tiered $5/3/1$ weights emphasizing contact and interaction regions~\cite{pan2026surgvista}, with newly uncovered background weighted at least $3$. The selection loss $\mathcal{L}_{\mathrm{sel}}$ applies cross-entropy over pixels where both $\widetilde{P}_t^{\mathrm{inst}}$ and $B$ are valid and have unequal $\ell_1$ errors to $P_t^*$. Hard depth resolution and reference selection stop gradients, whereas bilinear feature transport allows geometric errors to refine trajectories during Phase~3. We set $\lambda_{\mathrm{surf}}{=}0.10$, $\lambda_{\mathrm{sel}}{=}0.01$, and $\lambda_{\mathrm{pos}}{=}0.10$.

\modhead{Stage-2 Flow Matching and Distillation}
For Eq.~(14), given clean video latents $\mathbf{Z}_{\mathrm{clean}}$, Gaussian noise $\boldsymbol{\epsilon}{\sim}\mathcal{N}(\mathbf{0},\mathbf{I})$, $\tau{\sim}\mathcal{U}(0,1)$, and shifted noise level $\sigma{=}5\tau/(1{+}4\tau)$, we construct noisy future latents $(\mathbf{Z}_\sigma)_{\mathrm{fut}}{=}(1{-}\sigma)(\mathbf{Z}_{\mathrm{clean}})_{\mathrm{fut}}{+}\sigma\boldsymbol{\epsilon}_{\mathrm{fut}}$ while keeping $\mathbf{Z}_{\mathrm{his}}$ fixed, and optimize \[\mathcal{L}_{\mathrm{FM}}{=}\operatorname{Mean}_f\bigl((\mathbf{v}_{\mathrm{S}}{-}(\boldsymbol{\epsilon}_{\mathrm{fut}}{-}(\mathbf{Z}_{\mathrm{clean}})_{\mathrm{fut}}))^2\bigr)\]~\cite{lipman2023flow}. Distillation aligns control features~\cite{romero2015fitnets} and latent velocities over valid future entries, where $\operatorname{Mean}_f$, $\operatorname{Mean}_h$, and $\operatorname{Mean}_v$ average over valid future flow-matching, control-hint, and latent-velocity elements, respectively, and $\operatorname{sg}(\cdot)$ denotes stop-gradient:
\begin{align}
\label{eq:supp_lhint}
\mathcal{L}_{\mathrm{hint}}
&= \frac{1}{J}\sum_{j=1}^{J}
\frac{\operatorname{Mean}_h\!\left((\mathbf{H}_{\sigma,\mathrm{S}}^{(j)}-\operatorname{sg}(\mathbf{H}_{\sigma,\mathrm{T}}^{(j)}))^2\right)}
{\operatorname{Mean}_h\!\left((\operatorname{sg}(\mathbf{H}_{\sigma,\mathrm{T}}^{(j)}))^2\right)+\varepsilon}, \\
\label{eq:supp_lvel}
\mathcal{L}_{\mathrm{velocity}}
&= \operatorname{Mean}_v\!\left((\mathbf{v}_{\mathrm{S}}-\operatorname{sg}(\mathbf{v}_{\mathrm{T}}))^2\right),
\end{align}
where $\lambda_h{=}0.05$, $\lambda_v{=}0.5$, and $\varepsilon{=}10^{-8}$. At inference, we use 50-step FlowMatch Euler sampling with $\sigma$-shift~5 and CFG~1, clamping $\mathbf{Z}_{\mathrm{his}}$ at each step and restoring the five conditioning history frames after decoding. Table~\ref{tab:supp_hyperparams} summarizes all key hyperparameters.

\begin{table}[tbp]
\centering
\caption{\textbf{Summary of key hyperparameters} for Stage-1 geometry prediction and Stage-2 video generation.}
\label{tab:supp_hyperparams}
\small
\setlength{\tabcolsep}{0pt}
\renewcommand{\arraystretch}{1.12}
\begin{tabular*}{\linewidth}{@{\extracolsep{\fill}}lr@{}}
\toprule
\textbf{Hyperparameter} & \textbf{Value} \\
\midrule
\rowcolor{black!5}\multicolumn{2}{@{}l}{\textbf{Stage 1: Geometry Predictor (Secs.~3.2--3.3)}} \\
Frames ($N / T$) & 5 / 28 (short), 80 (long) @ 15~Hz \\
Loss weights ($\lambda_{\mathrm{surf}}, \lambda_{\mathrm{sel}}, \lambda_{\mathrm{pos}}$) & $0.10,\; 0.01,\; 0.10$ \\
Surface separations $\mathcal{R}$ & $\{1, 2, 4, 8\}$ (H \& V pairs) \\
Region weight $w_t(u)$ & Tiered $5 / 3 / 1$ (uncovered bg $\ge 3$) \\
Progressive phases & Traj.\ $\to$ frozen-traj.\ $\to$ end-to-end \\
Training budget & Main: 50K iters/dom.; Abl.: 15 ep. \\
Optimizer & AdamW (wd 0, grad clip 1.0) \\
Learning rate & Geo $10^{-4}$ / Traj.\ $10^{-5}$ \\
Batch / Precision & BS 12, BF16 \\
\midrule
\rowcolor{black!5}\multicolumn{2}{@{}l}{\textbf{Stage 2: Video Generator \& Distillation (Sec.~3.4)}} \\
Video backbone (frozen) & Wan2.2 TI2V-5B~\cite{wan2025wan} ($512{\times}288$) \\
Control / LoRA & 8L Transf.\ ($J{=}8$); LoRA $r{=}64, \alpha{=}32$ \\
Progressive phases & Control warmup $\to$ joint ctrl + LoRA \\
Training budget & Main: 50K iters/dom.; Abl.: 10K iters \\
Distillation ($\lambda_h, \lambda_v, \varepsilon$) & $0.05,\; 0.5,\; 10^{-8}$ \\
Optimizer & AdamW (wd 0, grad clip 1.0) \\
Learning rate & Ctrl $10^{-4}{\to}3{\times}10^{-5}$, LoRA $10^{-4}$ \\
Batch / Precision & BS 32, BF16 \\
Flow matching / Sampler & $\sigma{=}5\tau/(1{+}4\tau)$; Euler 50, shift 5, CFG 1 \\
\bottomrule
\end{tabular*}
\end{table}

\section{Experimental Protocols and Ablations}
\label{sec:supp_setup}

\subsection{Datasets, Splits, and Ablation Setup}
\label{sec:supp_datasets}

\modhead{Main Benchmark and Splits}
From Open-H-Embodiment~\cite{nelson2026open}, we use 29 surgical subsets comprising 79.05~hours of action-paired view-specific video across 33,846 sequences and 25,418 episodes. All data are partitioned at the episode level into training and test splits at an 8:1 ratio, assigning all views of each episode to the same split. As summarized in Table~\ref{tab:supp_datasets}, the Phantom domain covers 19 sources with 10,173 sequences and 39.02~hours of video, while the Ex vivo domain covers 10 sources with 23,673 sequences and 40.03~hours. For 80-frame long-window training, 33-frame short windows with 5 history and 28 future frames are co-trained with 85-frame long windows containing 5 history and 80 future frames, with aligned RGB, geometry, and actions. All test evaluations use continuous forward real recordings: Table~1 evaluates 1,280 test windows (640 per domain) across 655 held-out episodes from task subsets featuring instrument--tissue interaction; Tables~2--4 and Table~\ref{tab:supp_coarse_to_fine} evaluate 1,209 JHU IMERSE test windows; and Table~\ref{tab:supp_rollout} evaluates 32 Ex vivo episodes.

\modhead{Ablation Subset Selection}
To keep the controlled architecture and distillation ablations in Tables~2--4 computationally tractable while preserving task and domain diversity, we construct the 9,403-episode ablation suite from six representative task subsets of the JHU IMERSE collection in Open-H-Embodiment, spanning three Ex vivo and three Phantom surgical interaction primitives: cautery, fat dissection and retraction, porcine cholecystectomy, tissue lifting, needle handover, and suturing. Retained episodes pass unified sequence-validity and task-curation criteria (one 33-frame, 15~Hz physical window per episode with complete dual-arm kinematics and geometry preprocessing), spanning 16.95\% Phantom / 83.05\% Ex vivo by episode (vs.\ 21.8\% / 78.2\% in the full benchmark) and 23.5\% / 76.5\% across view-specific windows. Following the episode-level 8:1 train-to-test ratio of Sec.~4.1 with zero episode overlap, the ablation splits provide 9,636 training and 1,209 held-out test view windows.

\begin{table}[tbp]
\centering
\caption{\textbf{Dataset and evaluation summary} (episode-level 8:1 train/test splits) for the 29-source main benchmark and the 6-subset JHU IMERSE ablation suite.}
\label{tab:supp_datasets}
\small
\setlength{\tabcolsep}{0pt}
\renewcommand{\arraystretch}{1.12}
\begin{tabular*}{\linewidth}{@{\extracolsep{\fill}}lccr@{}}
\toprule
\textbf{Benchmark / Split} & \textbf{\# Src.} & \textbf{Total Ep.} & \textbf{Held-out Eval Set} \\
\midrule
\rowcolor{black!5}\multicolumn{4}{@{}l}{\textbf{Main Benchmark}} \\
Phantom (P) & 19 & 5,546 & 640 held-out windows \\
Ex vivo (E) & 10 & 19,872 & 640 held-out windows \\
\textbf{Total (Main)} & \textbf{29} & \textbf{25,418} & \textbf{1,280 windows (Table~1)} \\
\midrule
\rowcolor{black!5}\multicolumn{4}{@{}l}{\textbf{Ablation Suite}} \\
\textbf{Ablation (JHU)} & \textbf{6} & \textbf{9,403} & \textbf{1,209 windows (Tables~2--4)} \\
\bottomrule
\end{tabular*}
\end{table}

\begin{table}[tbp]
\centering
\caption{\textbf{Baseline setups on Open-H-Embodiment.} All six baselines are fine-tuned separately per domain (P and E) to convergence using their official architectures, adapted domain configurations, recommended optimization recipes, and standard inference sampling, and evaluated on identical future timestamps at $512{\times}288$.}
\label{tab:supp_baselines}
\small
\setlength{\tabcolsep}{5pt}
\renewcommand{\arraystretch}{1.12}
\begin{adjustbox}{max width=\linewidth}
\begin{tabularx}{\textwidth}{l l l X}
\toprule
\textbf{Method} & \textbf{Backbone \& Output Format} & \textbf{Conditioning \& Feedback} & \textbf{Per-Domain Adaptation} \\
\midrule
Ctrl-World~\cite{guo2026ctrl} & SVD ($320{\times}192$, 5\,Hz, 4\,f) & Replay memory + actions & 25K/dom.\ (BS 64, 1.60M), LR $10^{-5}$ \\
BWM~\cite{bwm2026bwm} & Wan2.2 5B ($512{\times}288$, 15\,Hz, 28\,f) & 5-f RGB hist.\ + actions & 20K/dom.\ (BS 64, 1.28M), LR $5{\times}10^{-5}$ \\
IRASim~\cite{zhu2025irasim} & XL/2 + SDXL ($512{\times}288$, 10\,Hz, 12\,f) & $t{=}0$ RGB + AdaLN actions & 30K/dom.\ (BS 64, 1.92M), LR $10^{-4}$ \\
Cosmos-H~\cite{nelson2026open} & Cosmos-H ($512{\times}288$, 10\,Hz, 12\,f) & $t{=}0$ RGB + actions & 20K/dom.\ EMA (BS 96, 1.92M), LR $3{\times}10^{-5}$ \\
SurgSora~\cite{chen2025surgsora} & Stage-2 ($256{\times}256$, 15\,Hz, 20\,f) & $t{=}0$ RGB + GT motion & 50K/dom.\ (BS 4, 200K), LR $2{\times}10^{-5}$ \\
RoboStereo~\cite{zhang2026robostereo} & Dual-tower ($512{\times}288$, 15\,Hz, 12\,f) & $t{=}0$ RGB + DA3 XYZ + act. & 20K/dom.\ (BS 32, 640K), LR $10^{-5}$ \\
\midrule
\rowcolor{black!5}\textbf{SurgReact (Ours)} & Wan2.2 5B ($512{\times}288$, 15\,Hz, 28/80\,f) & 5-f RGB + pred.\ pointmaps & 50K/dom.\ (BS 32, 1.60M, Table~\ref{tab:supp_hyperparams}) \\
\bottomrule
\end{tabularx}
\end{adjustbox}
\end{table}

\begin{table}[tbp]
\centering
\caption{\textbf{Controlled ablation matrix} summarizing the varied factors, fixed controls, and evaluation budget for each study in Secs.~4.3--4.4.}
\label{tab:supp_ablation_matrix}
\small
\setlength{\tabcolsep}{0pt}
\renewcommand{\arraystretch}{1.12}
\begin{tabularx}{\linewidth}{@{}p{15mm}XXX@{}}
\toprule
\textbf{Study} & \textbf{Varied Factor} & \textbf{Fixed Controls} & \textbf{Budget \& Test Split} \\
\midrule
Table~2 & ITP \& CR (\emph{Direct}, w/o ITP, w/o CR) & Same frozen 10K RGB teacher & 15 ep.\ geometry / 1,209 JHU windows \\
Table~3 & GFI (feature align) \& RS (learned selection) & Same frozen 10K RGB teacher & 15 ep.\ geometry / 1,209 JHU windows \\
Table~4 & FM-only vs.\ hint + velocity distillation & Same pred.\ pointmaps, Wan2.2 init & 10K video updates / 1,209 JHU windows \\
Fig.~5 & Real (8/batch) vs.\ Mixed (8+8/batch) & GR00T-N1.6-3B, VL frozen (excl.\ norm.), real labels & 20K policy updates / held-out windows (4 tasks) \\
\bottomrule
\end{tabularx}
\end{table}

\subsection{Baselines and Evaluation Metrics}
\label{sec:supp_baselines}

\modhead{Baseline Adaptation and Training Budgets}
All six baselines in Table~1 are adapted and fine-tuned separately per domain on identical episode splits while preserving each method's official architecture, conditioning design, recommended optimization recipe, and inference sampling until convergence. All methods are scored on identical target future timestamps at $512{\times}288$. Table~\ref{tab:supp_baselines} lists per-domain updates, batch size, total clip exposures, and learning rate. Because baselines differ in backbone, trainable ratio, learning rate, and batch size, forcing one fixed update count would under-train small-batch models or overfit large-batch ones. Compared with SurgReact's Stage-2 batch size of 32, Ctrl-World, BWM, and IRASim use default batch size 64 and Cosmos-H uses 96, so per-domain clip exposures for Cosmos-H at 1.92M, IRASim at 1.92M, and Ctrl-World at 1.60M match or exceed SurgReact. BWM at 1.28M with full-parameter fine-tuning and RoboStereo at 640K likewise follow the official training paradigms of their respective backbones and repositories to convergence.

We preserve each baseline's native architecture, conditioning interface, and adapted temporal configuration. Accordingly, SurgSora receives sparse motion cues extracted from ground-truth future video following its official setting, whereas all other methods condition solely on historical visuals and planned actions. On the Table~1 Short Window of 28 frames, BWM shares the same Wan2.2 TI2V-5B backbone, $512{\times}288$ resolution, 15~Hz frame rate, 5-frame RGB history, and single-chunk 28-frame forward pass as SurgReact. Under this matched inference setting, SurgReact achieves consistently higher instrument and tissue trajectory accuracy than BWM in both domains, consistent with the matched-budget geometry ablations in Tables~2--3. Cosmos-H~\cite{nelson2026open} benefits from large-scale surgical pretraining across 32 datasets and 9 robotic platforms, which may contribute to its strong short-window Ex vivo appearance scores; nevertheless, SurgReact achieves the highest instrument and tissue trajectory accuracy across all domains and horizons and overtakes Cosmos-H on PSNR, SSIM, and LPIPS over 21.33~s in Table~\ref{tab:supp_rollout}. Contemporaneous surgical world models including KVLR~\cite{li2026articulated}, SurgVista~\cite{pan2026surgvista}, and SurgCast~\cite{liu2026surgcast} either lacked public Open-H-compatible code and checkpoints prior to submission or target articulated-skeleton action representations, while we follow the SurgWorld-Bench~\cite{pan2026surgvista} interaction trajectory protocol.

\modhead{Evaluation Metrics}
\begin{itemize}
    \item \emph{Visual metrics:} Unpadded frames are resized via anti-aliased bilinear interpolation to $512{\times}288$. We compute L1 on $[0,1]$, PSNR in dB, SSIM~\cite{1284395} with an $11{\times}11$ window and $\sigma{=}1.5$, and AlexNet v0.1 LPIPS~\cite{zhang2018unreasonable}, averaged equally over future frames and windows.
    \item \emph{Trajectory metrics:} Following~\cite{NEURIPS2022_58168e8a,pan2026surgvista}, $\mathrm{AJ}$ and $\delta_{\mathrm{avg}}^{\mathrm{vis}}$ are evaluated purely in RGB space, independent of DA3. For Table~1 over 1.87 and 5.33~s, fixed $t{=}0$ queries use up to 4 points per instrument prioritizing the tip and up to 16 points in visible tissue regions. These points are tracked offline on 30~Hz ground truth and generated videos at their output frame rates with CoTracker3, then scored at aligned 5~Hz timestamps on $256{\times}256$-normalized coordinates over $\{1,2,4,8,16\}$~px thresholds on a 0--100 scale.
    \item \emph{Geometry metrics:} $E_{\mathrm{3D}}$ and $E_{\mathrm{n}}$ are evaluated on native half-resolution grids across all 28 future frames against valid DA3 pixels. $E_{\mathrm{3D}}$ is pointwise Euclidean error $\|\widehat{P}_t-P_t^*\|_2$ scaled by $10^3$, and $E_{\mathrm{n}}$ is oriented normal angular error $\operatorname{atan2}(\|\mathbf{n}\times\mathbf{n}^*\|{+}10^{-12},\mathbf{n}\cdot\mathbf{n}^*)$ in $[0^\circ,180^\circ]$ from four adjacent triangles, reported over shared Global, Instrument from the $t{=}0$ mask union, and Interaction contact-neighborhood ROI masks.
\end{itemize}

\subsection{Controlled Ablation and Policy Protocols}
\label{sec:supp_ablation_controls}

\begin{table}[tbp]
\centering
\caption{\textbf{Long-horizon 320-frame (21.33~s) autoregressive rollout benchmark on Ex vivo} (32 held-out episodes across 6 surgical sources, 106 aligned 5~Hz timestamps at $512{\times}288$). Among the seven compared models (top 7 rows), \textbf{bold} denotes best and \underline{underline} second-best.}
\label{tab:supp_rollout}
\small
\setlength{\tabcolsep}{3.5pt}
\renewcommand{\arraystretch}{1.12}
\begin{adjustbox}{max width=\linewidth}
\begin{tabularx}{\textwidth}{l c Y Y Y Y Y Y Y Y}
\toprule
\multirow{2}{*}{\textbf{Method}} & \multirow{2}{*}{\textbf{Rollout Schedule}} & \multicolumn{2}{c}{\textbf{Instr.\ Motion}} & \multicolumn{2}{c}{\textbf{Tissue Resp.}} & \multicolumn{4}{c}{\textbf{Visual Quality}} \\
\cmidrule(lr){3-4}\cmidrule(lr){5-6}\cmidrule(lr){7-10}
& & AJ$\uparrow$ & $\delta_{\mathrm{avg}}^{\mathrm{vis}}\uparrow$ & AJ$\uparrow$ & $\delta_{\mathrm{avg}}^{\mathrm{vis}}\uparrow$ & L1$\downarrow$ & PSNR$\uparrow$ & SSIM$\uparrow$ & LPIPS$\downarrow$ \\
\midrule
Ctrl-World~\cite{guo2026ctrl} & 5\,Hz ($4{\times}27$) & 28.14 & 45.94 & 48.00 & 67.55 & \textbf{0.0361} & 22.76 & \underline{0.8272} & \underline{0.1615} \\
BWM~\cite{bwm2026bwm} & 15\,Hz ($28{\times}12$) & 27.77 & 40.63 & \underline{53.68} & \underline{69.98} & 0.0492 & 21.23 & 0.7791 & 0.1670 \\
IRASim~\cite{zhu2025irasim} & 10\,Hz ($12{\times}18$) & 18.70 & 30.94 & 40.13 & 58.14 & 0.0546 & 20.60 & 0.7244 & 0.2057 \\
Cosmos-H~\cite{nelson2026open} & 10\,Hz ($12{\times}18$) & \underline{33.20} & \underline{48.70} & 43.05 & 61.85 & 0.0429 & \underline{23.08} & 0.8021 & 0.1646 \\
SurgSora~\cite{chen2025surgsora} & 15\,Hz ($20{\times}16$) & 12.78 & 24.13 & 44.64 & 63.18 & 0.2031 & 12.69 & 0.5333 & 0.5166 \\
RoboStereo~\cite{zhang2026robostereo} & 15\,Hz ($12{\times}27$) & 23.68 & 35.65 & 39.75 & 57.86 & 0.0552 & 21.04 & 0.7795 & 0.2054 \\
\rowcolor{black!5}\textbf{SurgReact (Ours)} & 15\,Hz ($80{\times}4$) & \textbf{38.58} & \textbf{52.34} & \textbf{55.73} & \textbf{72.52} & \underline{0.0405} & \textbf{23.69} & \textbf{0.8280} & \textbf{0.1564} \\
\midrule
\emph{SurgReact (Ref.\ Geo.)} & \emph{15\,Hz ($28{\times}12$)} & \emph{56.65} & \emph{75.17} & \emph{59.33} & \emph{75.65} & \emph{0.0292} & \emph{26.46} & \emph{0.8835} & \emph{0.1219} \\
\bottomrule
\end{tabularx}
\end{adjustbox}
\end{table}

\noindent Table~\ref{tab:supp_ablation_matrix} summarizes the varied factors and fixed controls for each study in Secs.~4.3--4.4, holding data splits, initialization, and sampling noise fixed within each comparison.

\modhead{Stage-1 Geometry Ablations}
In Tables~2--3, all variants train jointly on the JHU ablation training split of 9,636 view-specific windows for 15 epochs with matched sample exposure, optimizer, learning rates, and initialization. Trajectory-guided variants share the same pretrained trajectory module, which is frozen for 5 epochs before 10 epochs of end-to-end Stage-1 fine-tuning. In Table~2, \emph{Direct} removes trajectory prediction, reference selection, and residual refinement to regress XYZ directly; \emph{w/o ITP} anchors instruments at $t{=}0$ while retaining selection and refinement; and \emph{w/o CR} retains trajectory guidance and feature transport but regresses XYZ directly. In Table~3, removing GFI composes preliminary pointmaps prior to shared CNN encoding, dropping motion-aligned feature initialization, while removing RS replaces learned selection with fixed nearest-depth composition. To isolate geometry quality cleanly, every variant's predicted pointmaps are rendered by the same frozen 10K-update RGB teacher using 50-step Euler sampling with shift~5 and fixed noise over 28 future frames at 15~Hz on the 1,209 JHU test windows.

\modhead{Stage-2 Distillation and Cross-Table Comparison}
In Table~4, \emph{Student} with flow matching only and \emph{w/ Distillation} with $\mathcal{L}_{\mathrm{FM}}{+}0.05\mathcal{L}_{\mathrm{hint}}{+}0.5\mathcal{L}_{\mathrm{velocity}}$ start from identical Wan2.2 weights, a fresh control branch, newly initialized LoRA, and identical predicted pointmaps. Both train for 10,000 updates with control warmup followed by joint control and LoRA optimization while keeping base Wan2.2 frozen, and are compared against the frozen 10K-update \emph{Teacher} conditioned on DA3 reference pointmaps $P_{1:T}^*$, denoted ``GT pointmaps'' in Sec.~4.3. Tables~2--3 feed each variant's predicted pointmaps directly into this frozen reference-trained teacher to isolate geometry quality, yielding L1 $0.0256$, PSNR $25.44$, SSIM $0.8444$, and LPIPS $0.1025$ for full SurgReact. In contrast, Table~4 trains 10K-update students directly on predicted pointmaps to bridge the train--test geometry shift; \emph{w/ Distillation} improves over \emph{Student} on all four metrics and surpasses direct frozen-teacher rendering on L1 at $0.0219$, SSIM at $0.8626$, and LPIPS at $0.0864$, with comparable PSNR at $25.22$.

\modhead{Downstream Offline Policy Protocol}
In Fig.~5, we evaluate downstream policy post-training with GR00T-N1.6-3B~\cite{bjorck2025gr00t} on four single-view RGB tasks: \emph{needle handover}, \emph{needle passing}, \emph{knot tying}, and \emph{first clip application}. Per task, 300 real and 300 domain-matched SurgReact short windows yield $7{,}825$, $8{,}023$, $8{,}253$, and $6{,}621$ real frame--action pairs plus equal generated pairs. Each frame is paired with 50 future 20D dual-arm action steps at 30~Hz, with proprioceptive state masked and actions normalized using real training statistics. \emph{Real} samples 8 real frames per batch, whereas \emph{Mixed} augments the same 8 real frames with 8 generated frames per batch, matching real-data exposure and update steps. Both keep the vision--language backbone weights frozen except for their normalization layers, and fine-tune the projection layer and the 4-step flow-matching action head for 20,000 updates using AdamW with learning rate $3{\times}10^{-5}$, cosine decay, and $236{\times}414{\to}224{\times}392$ crop augmentation. Checkpoints are evaluated every 5K updates and tested on $32$, $196$, $254$, and $236$ held-out real view windows across the four tasks.

\section{Long-Horizon Autoregressive Rollout}
\label{sec:supp_rollout}

\subsection{Protocol and Quantitative Benchmark (21.33~s)}
\label{sec:supp_rollout_quant}

\noindent While Table~1 evaluates horizons up to 5.33~s over 80 frames, we further test long-term temporal stability through continuous autoregressive multi-chunk rollout over 320 future frames, corresponding to 21.33~s at 15~Hz, on the Ex vivo domain.

\modhead{Evaluation Cohort and Autoregressive Feedback}
From the episode-isolated Ex vivo test split, we evaluate all 32 episodes spanning at least 325 continuous frames (5 history + 320 future frames) at 15~Hz across six sources: 1 Hamlyn knot tying, 4 Hamlyn suturing~1, 16 Hamlyn suturing~2, 6 JHU NephFat, 3 JHU ARCade, and 2 UC Berkeley debridement. SurgReact rolls out four consecutive 80-frame chunks. At each chunk boundary, the last five generated RGB frames, cropped to the valid content region, are fed back as visual history for the next chunk; DA3 pointmaps, SAM~3.1 and SurgSLOT instrument masks, and 5-frame CoTracker3 keypoints are re-estimated automatically before predicting the next 80 pointmaps and RGB frames without any future ground-truth visuals. All baselines roll out autoregressively using the schedules in Table~\ref{tab:supp_rollout} while preserving their original feedback mechanisms from Table~\ref{tab:supp_baselines} (with SurgSora additionally receiving GT-derived sparse motion cues). Importantly, iterative multi-chunk rollout reflects the established long-horizon paradigm of these baselines rather than an artificial handicap: Ctrl-World~\cite{guo2026ctrl}, Cosmos-H~\cite{nelson2026open}, BWM~\cite{bwm2026bwm}, IRASim~\cite{zhu2025irasim}, and RoboStereo~\cite{zhang2026robostereo} all explicitly support sequential chunk-wise autoregression in their original formulations. In particular, models specifically architected around iterative short-chunk rollout to extend long horizons---most notably Ctrl-World (4-step 5~Hz rollout with evolving replay memory designed for 10--20\,s+ trajectories) and Cosmos-H (iterative 12-frame 10~Hz surgical rollouts)---operate within their intended multi-chunk rollout regime under this benchmark. By contrast, SurgReact's ability to sustain four coherent 80-frame chunks (only three inter-chunk handoffs) is supported by its lightweight Stage-1 geometric anchoring. We also include an upper-bound reference, \emph{SurgReact (Ref.\ Geo.)}, which runs the 28-frame reference-pointmap teacher over $28{\times}12$ chunks at 15~Hz while feeding back generated RGB history and conditioning on DA3 reference pointmaps from real future frames.

\modhead{Quantitative Findings}
All methods are evaluated on 106 aligned 5~Hz future timestamps from $0.2$~s to $21.2$~s at $512{\times}288$ using the same RGB metrics and continuous RGB-space CoTracker3 tracking, with denser queries than Table~1 over the 21.33-s span: 8 queries per instrument and up to 32 visible tissue queries per window. As reported in Table~\ref{tab:supp_rollout}, whereas Cosmos-H outperformed SurgReact on short-window Ex vivo appearance in Table~1, over 21.33~s SurgReact outperforms all six baselines on PSNR at 23.69~dB, SSIM at 0.8280, LPIPS at 0.1564, and all four instrument and tissue trajectory metrics, while ranking second on L1 at 0.0405 behind Ctrl-World (0.0361, consistent with its iterative replay-memory design for long-horizon appearance stability). In particular, SurgReact exceeds Cosmos-H by $+5.38$ Instrument AJ and $+12.68$ Tissue AJ, and exceeds BWM by $+10.81$ Instrument AJ and $+2.05$ Tissue AJ. These consistent gains over all six baselines across 21.33~s support explicit geometric anchoring for curbing long-horizon drift, while the remaining margin to \emph{SurgReact (Ref.\ Geo.)} at 26.46~dB PSNR, 56.65 Instrument AJ, and 59.33 Tissue AJ indicates further headroom when conditioned on oracle future geometry.

\subsection{Qualitative Multi-Chunk Rollout Comparisons}
\label{sec:supp_rollout_qual}

\noindent Figs.~\ref{fig:supp_rollout_w24} and~\ref{fig:supp_rollout_w28} visualize synchronized RGB frames and 3D pointmaps at chunk endpoints $t{=}0.00,\allowbreak 5.33,\allowbreak 10.67,\allowbreak 16.00$, and $21.33$~s on two representative Ex vivo episodes from JHU NephFat and JHU ARCade. Over the 21.33-s rollout, baselines accumulate visible drift: SurgSora degrades into high-contrast artifacts from 5.33~s, IRASim washes into blur, BWM drifts in color saturation and hallucinates tool structures, Cosmos-H warps tissue folds and shifts instrument tips, Ctrl-World lags in tool pose, and RoboStereo exhibits fragmented silhouettes and pointmap holes. In contrast, SurgReact maintains coherent instrument placement, cumulative tissue deformation, natural lighting, and smooth 3D pointmaps through 21.33~s.

\newcommand{\drawrolloutfig}[1]{\begin{adjustbox}{max width=\linewidth,max totalheight=.72\textheight}
\begin{tikzpicture}[x=1in, y=1in, font=\sffamily]
  \def\imgW{1.17}
  \def\imgH{0.658125}    \def\gapX{0.032}
  \def\gapY{0.030}
  \def\lblX{0.14}       \def\xStart{0.30}     \pgfmathsetmacro{\xColZ}{\xStart}
  \pgfmathsetmacro{\xColA}{\xColZ + \imgW + \gapX}
  \pgfmathsetmacro{\xColB}{\xColA + \imgW + \gapX}
  \pgfmathsetmacro{\xColC}{\xColB + \imgW + \gapX}
  \pgfmathsetmacro{\xColD}{\xColC + \imgW + \gapX}
  \pgfmathsetmacro{\xRight}{\xColD + \imgW}
  \pgfmathsetmacro{\rowStep}{\imgH + \gapY}

    \newcommand{\cellcover}[3]{    \begin{scope}
      \clip (##1, ##2 - 0.5*\imgH) rectangle (##1 + \imgW, ##2 + 0.5*\imgH);
      \node[inner sep=0pt, anchor=center] at (##1 + 0.5*\imgW, ##2) {        \includegraphics[width=\imgW in]{##3}      };
    \end{scope}
  }

    \def\yHdr{0.12}
  \foreach \xC/\tLab in {\xColZ/{$t{=}0.00$\,s}, \xColA/{$t{=}5.33$\,s}, \xColB/{$t{=}10.67$\,s}, \xColC/{$t{=}16.00$\,s}, \xColD/{$t{=}21.33$\,s}}{
    \node[anchor=base, font=\sffamily\small\bfseries, text=black!85] at (\xC + 0.5*\imgW, \yHdr) {\tLab};
  }

    \foreach \rIdx/\mDir/\mLabel in {
    0/reference/{Ground\\[-1.5pt]Truth},
    1/student_predicted/{\textbf{SurgReact}\\[-1.5pt]\textbf{(Ours)}},
    2/cosmos/{Cosmos-H},
    3/bwm/{BWM},
    4/ctrl/{Ctrl-World},
    5/robostereo/{RoboStereo},
    6/irasim/{IRASim},
    7/surgsora/{SurgSora$^\dagger$}
  }{
    \pgfmathsetmacro{\yR}{-\rIdx*\rowStep - 0.5*\imgH}
    \node[rotate=90, align=center, font=\sffamily\small, text=black!85] at (\lblX, \yR) {\mLabel};
    \cellcover{\xColZ}{\yR}{fig/rollout/#1/RGB/reference/f000.png}
    \cellcover{\xColA}{\yR}{fig/rollout/#1/RGB/\mDir/f080.png}
    \cellcover{\xColB}{\yR}{fig/rollout/#1/RGB/\mDir/f160.png}
    \cellcover{\xColC}{\yR}{fig/rollout/#1/RGB/\mDir/f240.png}
    \cellcover{\xColD}{\yR}{fig/rollout/#1/RGB/\mDir/f320.png}
  }

  \pgfmathsetmacro{\yRgbBot}{-8*\imgH - 7*\gapY}

    \def\secGap{0.20}
  \pgfmathsetmacro{\yDiv}{\yRgbBot - 0.52*\secGap}
  \node[anchor=center, font=\sffamily\small\bfseries, text=black!80] at (0.5*\xColZ + 0.5*\xRight, \yDiv) {Pointmap (XYZ $\to$ RGB)};

    \pgfmathsetmacro{\yPmTop}{\yRgbBot - \secGap}
  \foreach \rIdx/\pDir/\pLabel in {
    0/teacher_gt/{Ref.\\[-1.5pt]Pointmap},
    1/student_predicted/{\textbf{SurgReact}\\[-1.5pt]\textbf{(Ours)}},
    2/robostereo/{RoboStereo}
  }{
    \pgfmathsetmacro{\yP}{\yPmTop - \rIdx*\rowStep - 0.5*\imgH}
    \node[rotate=90, align=center, font=\sffamily\small, text=black!85] at (\lblX, \yP) {\pLabel};
    \cellcover{\xColZ}{\yP}{fig/rollout/#1/Pointmap/\pDir/f000.png}
    \cellcover{\xColA}{\yP}{fig/rollout/#1/Pointmap/\pDir/f080.png}
    \cellcover{\xColB}{\yP}{fig/rollout/#1/Pointmap/\pDir/f160.png}
    \cellcover{\xColC}{\yP}{fig/rollout/#1/Pointmap/\pDir/f240.png}
    \cellcover{\xColD}{\yP}{fig/rollout/#1/Pointmap/\pDir/f320.png}
  }
\end{tikzpicture}
\end{adjustbox}}

\begin{figure}[p]
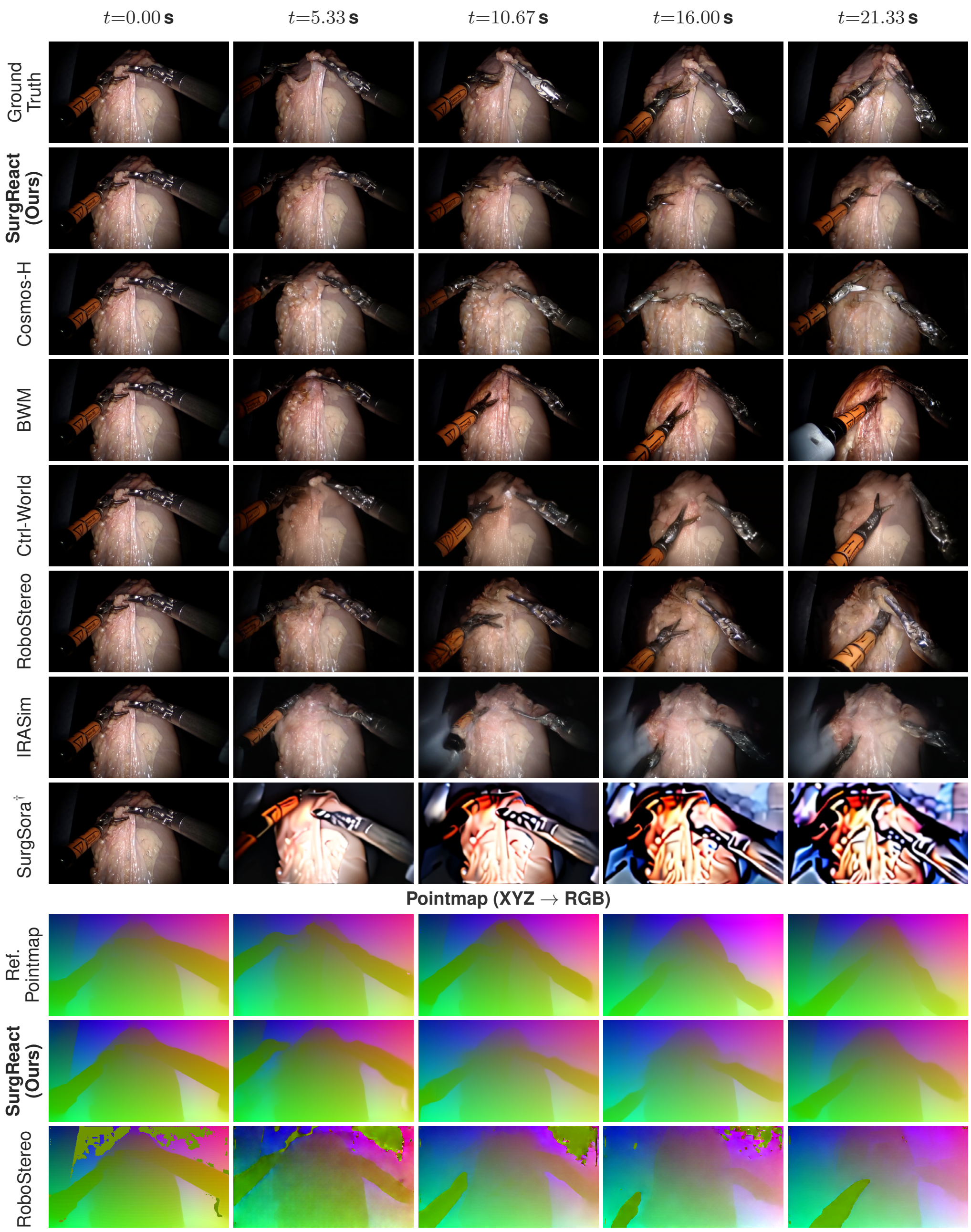

\centering
\drawrolloutfig{W24}
\vspace{1mm}
\caption{\textbf{Qualitative 21.33-s autoregressive rollout on an ex vivo tissue-retraction scenario.} Top: RGB rollout comparing Ground Truth against all seven models starting from the shared $t{=}0.00$\,s observation ($^\dagger$SurgSora uses GT-derived sparse motion cues). Bottom: synchronized 3D pointmaps (XYZ $\to$ RGB in shared per-window 1st--99th percentile bounds). SurgReact maintains coherent dual-arm motion, tissue retraction, natural lighting, and smooth 3D geometry across all four 80-frame chunks.}
\label{fig:supp_rollout_w24}
\end{figure}

\begin{figure}[p]
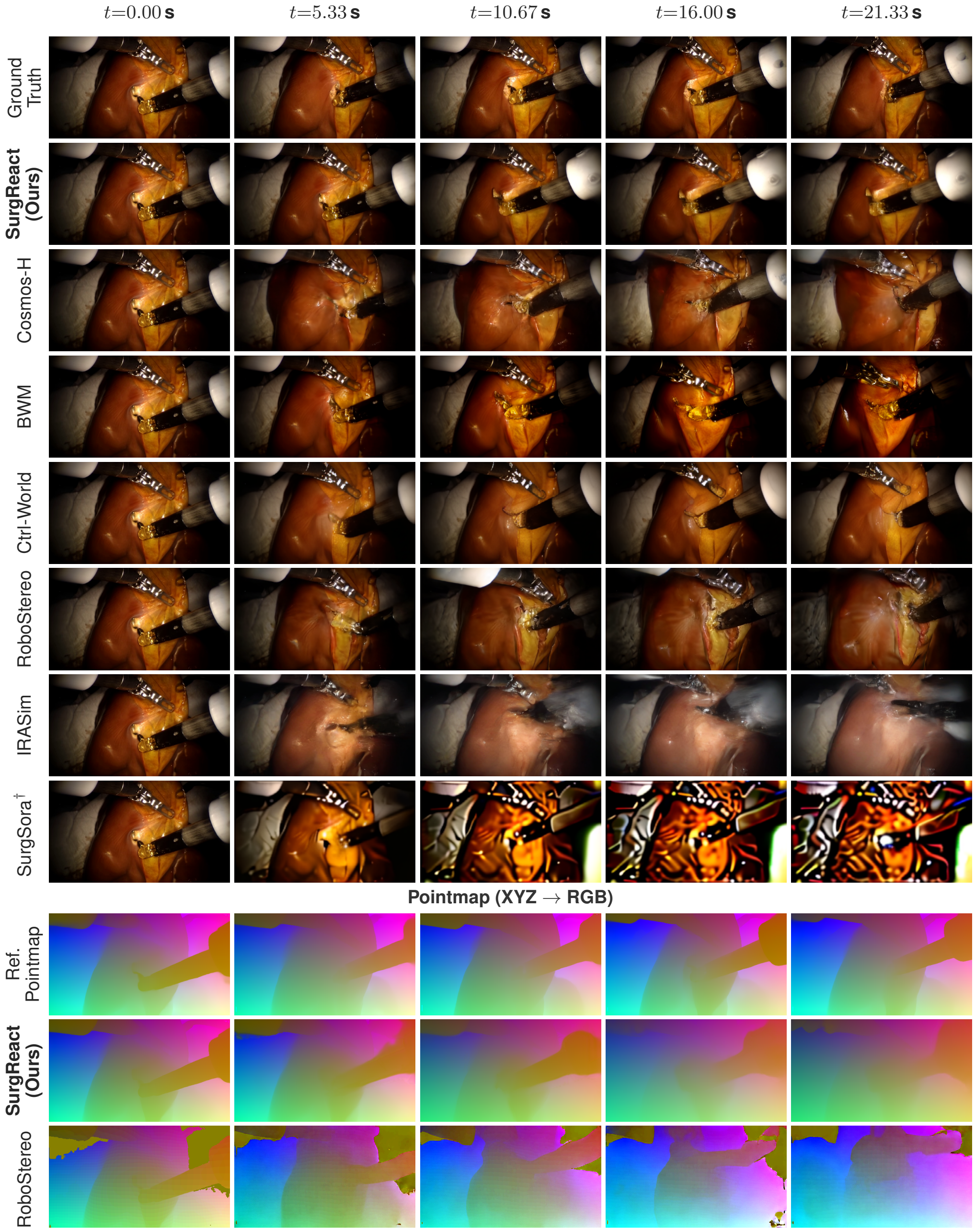

\centering
\drawrolloutfig{W28}
\vspace{1mm}
\caption{\textbf{Qualitative 21.33-s autoregressive rollout on an ex vivo cholecystectomy scenario.} Top: RGB rollout comparing Ground Truth against all seven models from the shared $t{=}0.00$\,s frame ($^\dagger$SurgSora uses GT-derived sparse motion cues). Bottom: synchronized 3D pointmaps (XYZ $\to$ RGB). SurgReact preserves crisp metallic contours, tissue folds, and continuous 3D scene geometry through 21.33~s.}
\label{fig:supp_rollout_w28}
\end{figure}

\section{Diagnostics and Qualitative Results}
\label{sec:supp_diagnostics}

\subsection{Quantitative Geometry vs.\ RoboStereo (Fig.~4)}
\label{sec:supp_robostereo_geo}

\noindent Complementing Fig.~4, Table~\ref{tab:supp_robostereo_geo} compares 28-frame pointmap accuracy between our Stage-1 predictor, matching the SurgReact checkpoint in Tables~2--3 and Table~\ref{tab:supp_coarse_to_fine}, and RoboStereo~\cite{zhang2026robostereo} over $12{+}12{+}4$ native RGB/XYZ chunks on the JHU test split with identical DA3 references and masks. Under this decoupled formulation, SurgReact achieves $3.3\times$--$4.7\times$ lower 3D coordinate error $E_{\mathrm{3D}}$ across all three regions and $2.7\times$--$4.5\times$ lower surface-normal error $E_{\mathrm{n}}$ compared with joint RGB--pointmap diffusion in RoboStereo.

\begin{table}[ht!]
\centering
\caption{\textbf{Quantitative pointmap comparison} against RoboStereo~\cite{zhang2026robostereo} (complementing Fig.~4 of the main paper).}
\label{tab:supp_robostereo_geo}
\small
\setlength{\tabcolsep}{2.5pt}
\renewcommand{\arraystretch}{1.12}
\begin{adjustbox}{max width=\linewidth}
\begin{tabularx}{\linewidth}{l Y Y Y Y Y Y}
\toprule
\multirow{2}{*}{\textbf{Method}} & \multicolumn{2}{c}{\textbf{Global}} & \multicolumn{2}{c}{\textbf{Instrument}} & \multicolumn{2}{c}{\textbf{Interaction}} \\
\cmidrule(lr){2-3} \cmidrule(lr){4-5} \cmidrule(lr){6-7}
& $E_{\mathrm{3D}}\downarrow$ & $E_{\mathrm{n}}\downarrow$ & $E_{\mathrm{3D}}\downarrow$ & $E_{\mathrm{n}}\downarrow$ & $E_{\mathrm{3D}}\downarrow$ & $E_{\mathrm{n}}\downarrow$ \\
\midrule
RoboStereo~\cite{zhang2026robostereo} & 104.30 & 74.61 & 92.58 & 77.37 & 71.88 & 74.58 \\
\rowcolor{black!5}\textbf{SurgReact (Ours)} & \textbf{22.37} & \textbf{16.72} & \textbf{27.84} & \textbf{28.87} & \textbf{20.31} & \textbf{18.88} \\
\bottomrule
\end{tabularx}
\end{adjustbox}
\end{table}

\subsection{Stage-Wise Refinement and Inference Efficiency}
\label{sec:supp_coarse_runtime}

\begin{figure}[p]
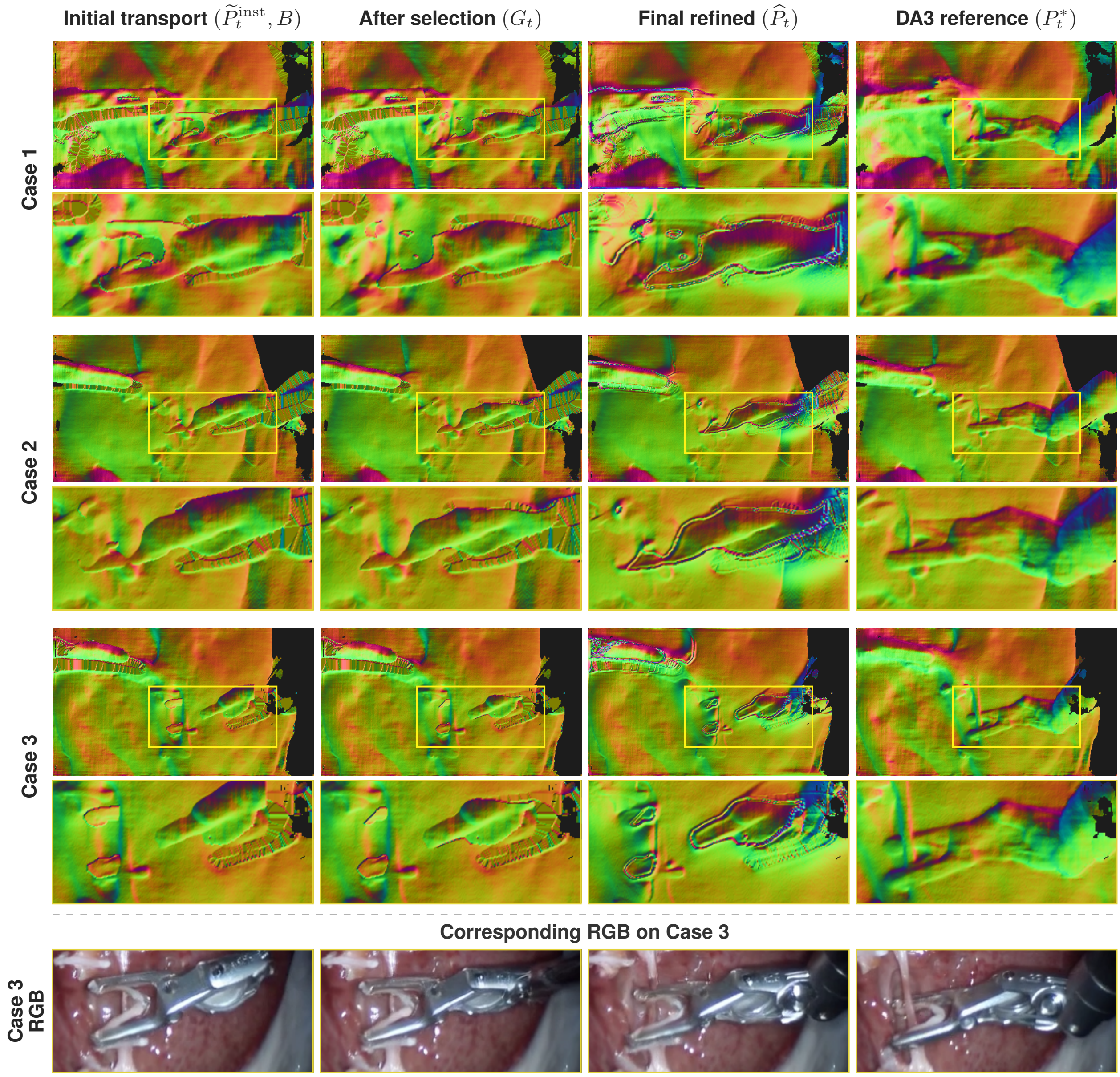

\centering
\begin{adjustbox}{max width=\linewidth,max totalheight=.72\textheight}
\begin{tikzpicture}[x=1in, y=1in, font=\sffamily]
  \def\imgW{1.56}
  \def\fullH{0.8817}    \def\zoomH{0.7372}    \def\gapX{0.045}
  \def\pairGap{0.026}   \def\caseGap{0.115}   \def\lblX{0.12}
  \def\xStart{0.26}
  \pgfmathsetmacro{\caseStep}{\fullH + \pairGap + \zoomH + \caseGap}

    \foreach \k/\hLab in {
    0/{Initial transport $(\widetilde{P}_t^{\mathrm{inst}}, B)$},
    1/{After selection $(G_t)$},
    2/{Final refined $(\widehat{P}_t)$},
    3/{DA3 reference $(P_t^*)$}
  }{
    \pgfmathsetmacro{\xC}{\xStart + \k*(\imgW + \gapX)}
    \node[anchor=base, font=\sffamily\small\bfseries, text=black!85] at (\xC + 0.5*\imgW, 0.10) {\hLab};
  }

    \foreach \ci in {1,2,3}{
    \pgfmathsetmacro{\yTop}{-(\ci - 1)*\caseStep}
    \pgfmathsetmacro{\yFullBot}{\yTop - \fullH}
    \pgfmathsetmacro{\yZoomTop}{\yFullBot - \pairGap}
    \pgfmathsetmacro{\yZoomBot}{\yZoomTop - \zoomH}
    \pgfmathsetmacro{\yCaseMid}{0.5*(\yTop + \yZoomBot)}

        \node[rotate=90, font=\sffamily\small\bfseries, text=black!85] at (\lblX, \yCaseMid) {Case~\ci};

    \foreach \k in {0,1,2,3}{
      \pgfmathsetmacro{\xL}{\xStart + \k*(\imgW + \gapX)}
            \node[inner sep=0pt, anchor=south west] at (\xL, \yFullBot) {        \includegraphics[width=\imgW in, height=\fullH in]{fig/coarse_to_fine/c\ci_p\k.png}      };
            \draw[yellow!95!white, line width=0.75pt]
        (\xL + 0.37*\imgW, \yFullBot + 0.20*\fullH) rectangle
        (\xL + 0.86*\imgW, \yFullBot + 0.61*\fullH);
            \node[inner sep=0pt, anchor=south west] at (\xL, \yZoomBot) {        \includegraphics[width=\imgW in, height=\zoomH in]{fig/coarse_to_fine/c\ci_z\k.png}      };
      \draw[yellow!85!black, line width=0.65pt]
        (\xL, \yZoomBot) rectangle (\xL + \imgW, \yZoomTop);
    }
  }

    \pgfmathsetmacro{\yCThreeBot}{-2*\caseStep - \fullH - \pairGap - \zoomH}
  \def\rgbDivGap{0.23}
  \pgfmathsetmacro{\yDivLine}{\yCThreeBot - 0.28*\rgbDivGap}
  \pgfmathsetmacro{\yDivTxt}{\yCThreeBot - 0.78*\rgbDivGap}
  \pgfmathsetmacro{\yRgbTop}{\yCThreeBot - \rgbDivGap - 0.045}
  \pgfmathsetmacro{\yRgbBot}{\yRgbTop - \zoomH}
  \pgfmathsetmacro{\xRightEdge}{\xStart + 4*\imgW + 3*\gapX}

  \draw[black!30, dashed, line width=0.45pt] (\xStart, \yDivLine) -- (\xRightEdge, \yDivLine);
  \node[anchor=center, font=\sffamily\small\bfseries, text=black!80]
    at (0.5*\xStart + 0.5*\xRightEdge, \yDivTxt)
    {Corresponding RGB on Case~3};
  \node[rotate=90, align=center, font=\sffamily\small\bfseries, text=black!85]
    at (\lblX - 0.02, 0.5*\yRgbTop + 0.5*\yRgbBot) {Case~3\\[-1.5pt]RGB};

  \foreach \k in {0,1,2,3}{
    \pgfmathsetmacro{\xL}{\xStart + \k*(\imgW + \gapX)}
    \node[inner sep=0pt, anchor=south west] at (\xL, \yRgbBot) {      \includegraphics[width=\imgW in, height=\zoomH in]{fig/coarse_to_fine/c3_rgb_z\k.png}    };
    \draw[yellow!85!black, line width=0.65pt]
      (\xL, \yRgbBot) rectangle (\xL + \imgW, \yRgbTop);
  }
\end{tikzpicture}
\end{adjustbox}
\vspace{1mm}
\caption{\textbf{Coarse-to-refined 3D geometry progression} ($\widetilde{P}_t^{\mathrm{inst}} \to G_t \to \widehat{P}_t \to P_t^*$) on three held-out surgical cases with pronounced wrist rotation ($\ge 25^\circ$) or jaw articulation, visualized as surface-normal maps ($\mathrm{RGB}=(\mathbf{n}{+}1){\times}127.5$; dark gray denotes invalid reference pixels). Each yellow-framed wrist/jaw crop aligns flush beneath its full frame, showing how selection and residual refinement recover articulated instrument surfaces and surrounding tissue geometry. The bottom row shows the corresponding RGB crops on Case~3 rendered by the shared frozen video generator alongside Ground Truth, confirming that stage-wise geometric refinement translates into coherent instrument structure and appearance.}
\label{fig:supp_coarse_to_fine}
\end{figure}

\modhead{Coarse Transport to Refined Geometry}
Single-keypoint surface transport $\widetilde{P}_t^{\mathrm{inst}}$ in Eq.~(3) provides a coarse surface-transport anchor, while $\mathcal{C}_{\mathrm{sel}}$ and $\mathcal{F}_{\mathrm{refine}}$ recover wrist rotation, jaw articulation, and soft-tissue deformation. Across all 1,209 JHU test windows in Table~\ref{tab:supp_coarse_to_fine}, learned reference selection $G_t$ improves over nearest-depth composition, and selection plus coordinate refinement $\widehat{P}_t$ reduces $E_{\mathrm{3D}}$ relative to initial transport by 27.3\% Global, 35.6\% Instrument, and 26.6\% Interaction, improving 87.3\%, 89.5\%, and 84.7\% of windows, respectively. While initial surface transport retains sharp static local normals at the expense of large global displacement, selection and refinement correct these positional errors, reducing Interaction $E_{\mathrm{3D}}$ from $27.66$ to $20.31$ while also lowering $E_{\mathrm{n}}$ from $19.09^\circ$ to $18.88^\circ$. Fig.~\ref{fig:supp_coarse_to_fine} illustrates this progression on three held-out cases with pronounced wrist rotation of at least $25^\circ$ or jaw articulation. The bottom row of Fig.~\ref{fig:supp_coarse_to_fine} further renders the corresponding Case~3 region with the shared frozen video generator to illustrate how these geometric stages propagate to RGB synthesis. Starting from initial surface transport with incomplete shaft geometry and open jaws, learned reference selection $G_t$ improves shaft continuity along the wrist, and multiscale coordinate refinement $\widehat{P}_t$ recovers wrist posture and jaw closure toward the central tissue, closely matching Ground Truth and confirming that Stage-1 geometric improvements directly support coherent Stage-2 appearance.

\begin{table}[tbp]
\centering
\caption{\textbf{Stage-wise geometric error reduction} inside SurgReact (single-pass inference diagnostic, 1,209 JHU test windows; $\widehat{P}_t{=}G_t{+}\Delta P_t$; $E_{\mathrm{3D}}$ in $\times 10^3$, $E_{\mathrm{n}}$ in degrees).}
\label{tab:supp_coarse_to_fine}
\small
\setlength{\tabcolsep}{2.5pt}
\renewcommand{\arraystretch}{1.12}
\begin{adjustbox}{max width=\linewidth}
\begin{tabularx}{\linewidth}{l Y Y Y Y Y Y}
\toprule
\multirow{2}{*}{\textbf{Geometry Stage}} & \multicolumn{2}{c}{\textbf{Global}} & \multicolumn{2}{c}{\textbf{Instrument}} & \multicolumn{2}{c}{\textbf{Interaction}} \\
\cmidrule(lr){2-3} \cmidrule(lr){4-5} \cmidrule(lr){6-7}
& $E_{\mathrm{3D}}\downarrow$ & $E_{\mathrm{n}}\downarrow$ & $E_{\mathrm{3D}}\downarrow$ & $E_{\mathrm{n}}\downarrow$ & $E_{\mathrm{3D}}\downarrow$ & $E_{\mathrm{n}}\downarrow$ \\
\midrule
\addlinespace[1.5pt]
Initial transp.\ $(\widetilde{P}_t^{\mathrm{inst}}, B)$ & 30.78 & 16.17 & 43.26 & \textbf{27.63} & 27.66 & 19.09 \\
After selection $(G_t)$ & 30.54 & \textbf{16.14} & 43.04 & 27.90 & 27.35 & 19.02 \\
\rowcolor{black!5}\textbf{Final refined} $(\widehat{P}_t)$ & \textbf{22.37} & 16.72 & \textbf{27.84} & 28.87 & \textbf{20.31} & \textbf{18.88} \\
\bottomrule
\end{tabularx}
\end{adjustbox}
\end{table}

\modhead{Per-Stage Inference Breakdown}
Table~\ref{tab:supp_runtime} reports peak VRAM, latency, and time share for a 28-frame $512{\times}288$ clip at BS~1 in BF16 with 50-step Euler sampling, averaged over 10 test clips after 2 warmups on a 24~GB NVIDIA RTX~4090 with sequential execution and CUDA synchronization. History preprocessing runs once per chunk on its five conditioning frames and amortizes over 80-frame chunks, whereas Stage-1 geometry takes only 0.21~s, or 1.63\% of runtime, and 2.12~GB VRAM. Total single-chunk latency of 12.73~s at 15.71~GB peak VRAM is dominated by Stage-2 video diffusion at 77.23\%. On the same RTX~4090 and 10 clips covering at least 1.87~s, multi-chunk baselines round up to 2.40--2.67~s. SurgReact's 12.73~s latency, or 6.82~s per video second, matches 2D action-conditioned baselines such as Cosmos-H at 12.31~s and Ctrl-World at 11.70~s, while running $3.9\times$ faster in wall-clock time and $3.0\times$ faster per video second than RoboStereo at 49.53~s and 20.64~s/s.

\begin{table}[tbp]
\centering
\caption{\textbf{Per-stage inference cost breakdown of SurgReact} (28-frame clip at $512{\times}288$, BS=1, single NVIDIA RTX~4090, 10-clip mean).}
\label{tab:supp_runtime}
\small
\setlength{\tabcolsep}{3pt}
\renewcommand{\arraystretch}{1.12}
\begin{adjustbox}{max width=\linewidth}
\begin{tabularx}{\linewidth}{l Y Y Y}
\toprule
\textbf{Pipeline Stage} & \textbf{VRAM (GB)} & \textbf{Latency (s)} & \textbf{Share (\%)} \\
\midrule
History preprocessing & 12.14 & 2.69 & 21.14 \\
Stage-1 geometry predictor & 2.12 & 0.21 & 1.63 \\
Stage-2 video generator & 15.71 & 9.83 & 77.23 \\
\midrule
\rowcolor{black!5}\textbf{Total} & \textbf{15.71} & \textbf{12.73} & \textbf{100.00} \\
\bottomrule
\end{tabularx}
\end{adjustbox}
\end{table}

\subsection{Qualitative Component Ablation (Table~2)}
\label{sec:supp_ablation_qual}

\begin{figure}[tbp]
\centering
\begin{adjustbox}{max width=.91\linewidth,max totalheight=.72\textheight}
\begin{tikzpicture}[x=1in, y=1in, font=\sffamily]
  \def\imgW{1.06}
  \def\imgH{0.59625}     \def\zoomH{0.5300}     \def\gapX{0.032}
  \def\startGap{0.088}   \def\gapY{0.026}
  \def\caseGap{0.22}
  \def\lblX{0.12}
  \def\xStart{0.26}

  \pgfmathsetmacro{\xColZ}{\xStart}
  \pgfmathsetmacro{\xColA}{\xColZ + \imgW + \startGap}
  \pgfmathsetmacro{\xColB}{\xColA + \imgW + \gapX}
  \pgfmathsetmacro{\xColC}{\xColB + \imgW + \gapX}
  \pgfmathsetmacro{\xColD}{\xColC + \imgW + \gapX}
  \pgfmathsetmacro{\xColE}{\xColD + \imgW + \gapX}
  \pgfmathsetmacro{\xRight}{\xColE + \imgW}
  \pgfmathsetmacro{\xSep}{\xColZ + \imgW + 0.5*\startGap}
  \pgfmathsetmacro{\xMid}{0.5*\xStart + 0.5*\xRight}

    \foreach \xC/\hLab in {
    \xColZ/{Start},
    \xColA/{Direct},
    \xColB/{w/o ITP},
    \xColC/{w/o CR},
    \xColD/{\textbf{SurgReact (Ours)}},
    \xColE/{GT / DA3 Ref.}
  }{
    \node[anchor=base, font=\sffamily\small\bfseries, text=black!85] at (\xC + 0.5*\imgW, 0.08) {\hLab};
  }

    \definecolor{v0}{HTML}{440154}
  \definecolor{v1}{HTML}{3B528B}
  \definecolor{v2}{HTML}{21918C}
  \definecolor{v3}{HTML}{5EC962}
  \definecolor{v4}{HTML}{FDE725}

      \newcommand{\drawablationcase}[7]{    \pgfmathsetmacro{\yBan}{#1 - 0.088}
    \pgfmathsetmacro{\yRgbTop}{#1 - 0.17}
    \pgfmathsetmacro{\yRgbBot}{\yRgbTop - \imgH}
    \pgfmathsetmacro{\yDepTop}{\yRgbBot - \gapY}
    \pgfmathsetmacro{\yDepBot}{\yDepTop - \imgH}
    \pgfmathsetmacro{\yZTop}{\yDepBot - \gapY}
    \pgfmathsetmacro{\yZBot}{\yZTop - \zoomH}

        \fill[black!5] (\xStart, \yRgbTop + 0.015) rectangle (\xRight, #1);
    \node[anchor=west, font=\sffamily\small\bfseries, text=black!85] at (\xStart + 0.05, \yBan) {#3};

        \draw[black!30, dashed, line width=0.45pt] (\xSep, \yZBot) -- (\xSep, \yRgbTop);

        \node[rotate=90, align=center, font=\sffamily\small\bfseries, text=black!85]
      at (\lblX, 0.5*\yRgbTop + 0.5*\yRgbBot) {RGB};
    \node[rotate=90, align=center, font=\sffamily\small\bfseries, text=black!85]
      at (\lblX, 0.5*\yDepTop + 0.5*\yDepBot) {Depth};
    \node[rotate=90, align=center, font=\sffamily\small\bfseries, text=black!85]
      at (\lblX, 0.5*\yZTop + 0.5*\yZBot) {Depth\\[-1.5pt]Zoom};

        \node[inner sep=0pt, anchor=south west] at (\xColZ, \yRgbBot) {      \includegraphics[width=\imgW in, height=\imgH in]{fig/ablation_table2/#2_start_rgb.png}    };
    \node[inner sep=0pt, anchor=south west] at (\xColZ, \yDepBot) {      \includegraphics[width=\imgW in, height=\imgH in]{fig/ablation_table2/#2_start_depth.png}    };
    \draw[yellow!95!white, line width=0.7pt]
      (\xColZ + #4*\imgW, \yDepBot + #6*\imgH) rectangle
      (\xColZ + #5*\imgW, \yDepBot + #7*\imgH);
    \node[inner sep=0pt, anchor=south west] at (\xColZ, \yZBot) {      \includegraphics[width=\imgW in, height=\zoomH in]{fig/ablation_table2/#2_start_depth_z.png}    };
    \draw[yellow!85!black, line width=0.65pt]
      (\xColZ, \yZBot) rectangle (\xColZ + \imgW, \yZTop);

        \foreach \xC/\rKey/\dKey in {      \xColA/direct/direct,      \xColB/wo_itp/wo_itp,      \xColC/wo_cr/wo_cr,      \xColD/full/full,      \xColE/gt/ref}{
            \node[inner sep=0pt, anchor=south west] at (\xC, \yRgbBot) {        \includegraphics[width=\imgW in, height=\imgH in]{fig/ablation_table2/#2_rgb_\rKey.png}      };
            \node[inner sep=0pt, anchor=south west] at (\xC, \yDepBot) {        \includegraphics[width=\imgW in, height=\imgH in]{fig/ablation_table2/#2_depth_\dKey.png}      };
      \draw[yellow!95!white, line width=0.7pt]
        (\xC + #4*\imgW, \yDepBot + #6*\imgH) rectangle
        (\xC + #5*\imgW, \yDepBot + #7*\imgH);
            \node[inner sep=0pt, anchor=south west] at (\xC, \yZBot) {        \includegraphics[width=\imgW in, height=\zoomH in]{fig/ablation_table2/#2_depth_z_\dKey.png}      };
      \draw[yellow!85!black, line width=0.65pt]
        (\xC, \yZBot) rectangle (\xC + \imgW, \yZTop);
    }
  }

    \drawablationcase{0.00}{c07}{Case~A: Phantom Suturing}{0.5703}{0.9609}{0.3403}{0.6875}

    \pgfmathsetmacro{\yCaseH}{0.17 + 2*\imgH + 2*\gapY + \zoomH}
  \pgfmathsetmacro{\yCaseB}{-\yCaseH - \caseGap}
  \drawablationcase{\yCaseB}{c08}{Case~B: Ex Vivo Porcine Cholecystectomy}{0.3203}{0.7500}{0.0833}{0.4653}

    \pgfmathsetmacro{\yBottom}{\yCaseB - \yCaseH}
  \pgfmathsetmacro{\yTime}{\yBottom - 0.135}
  \foreach \xC/\tLab in {
    \xColZ/{$t{=}0.07$\,s},
    \xColA/{$t{=}1.87$\,s},
    \xColB/{$t{=}1.87$\,s},
    \xColC/{$t{=}1.87$\,s},
    \xColD/{$t{=}1.87$\,s},
    \xColE/{$t{=}1.87$\,s}
  }{
    \node[anchor=base, font=\sffamily\small, text=black!75] at (\xC + 0.5*\imgW, \yTime) {\tLab};
  }

    \def\barW{1.60}
  \def\barH{0.068}
  \pgfmathsetmacro{\barQ}{0.25*\barW}
  \pgfmathsetmacro{\barL}{\xMid - 0.5*\barW}
  \pgfmathsetmacro{\yBarMid}{\yTime - 0.145}
  \pgfmathsetmacro{\barB}{\yBarMid - 0.5*\barH}
  \pgfmathsetmacro{\barT}{\yBarMid + 0.5*\barH}
  \shade[left color=v0, right color=v1] (\barL, \barB) rectangle (\barL + \barQ, \barT);
  \shade[left color=v1, right color=v2] (\barL + \barQ, \barB) rectangle (\barL + 2*\barQ, \barT);
  \shade[left color=v2, right color=v3] (\barL + 2*\barQ, \barB) rectangle (\barL + 3*\barQ, \barT);
  \shade[left color=v3, right color=v4] (\barL + 3*\barQ, \barB) rectangle (\barL + 4*\barQ, \barT);
  \draw[black!35, line width=0.35pt] (\barL, \barB) rectangle (\barL + \barW, \barT);
  \node[anchor=east, font=\sffamily\small, text=black!75, inner sep=0pt] at (\barL - 0.06, \yBarMid) {Depth:\enskip Far};
  \node[anchor=west, font=\sffamily\small, text=black!75, inner sep=0pt] at (\barL + \barW + 0.06, \yBarMid) {Near};
\end{tikzpicture}
\end{adjustbox}
\vspace{1mm}
\caption{\textbf{Qualitative component ablation of various variants.}}
\label{fig:supp_ablation_t2}
\end{figure}
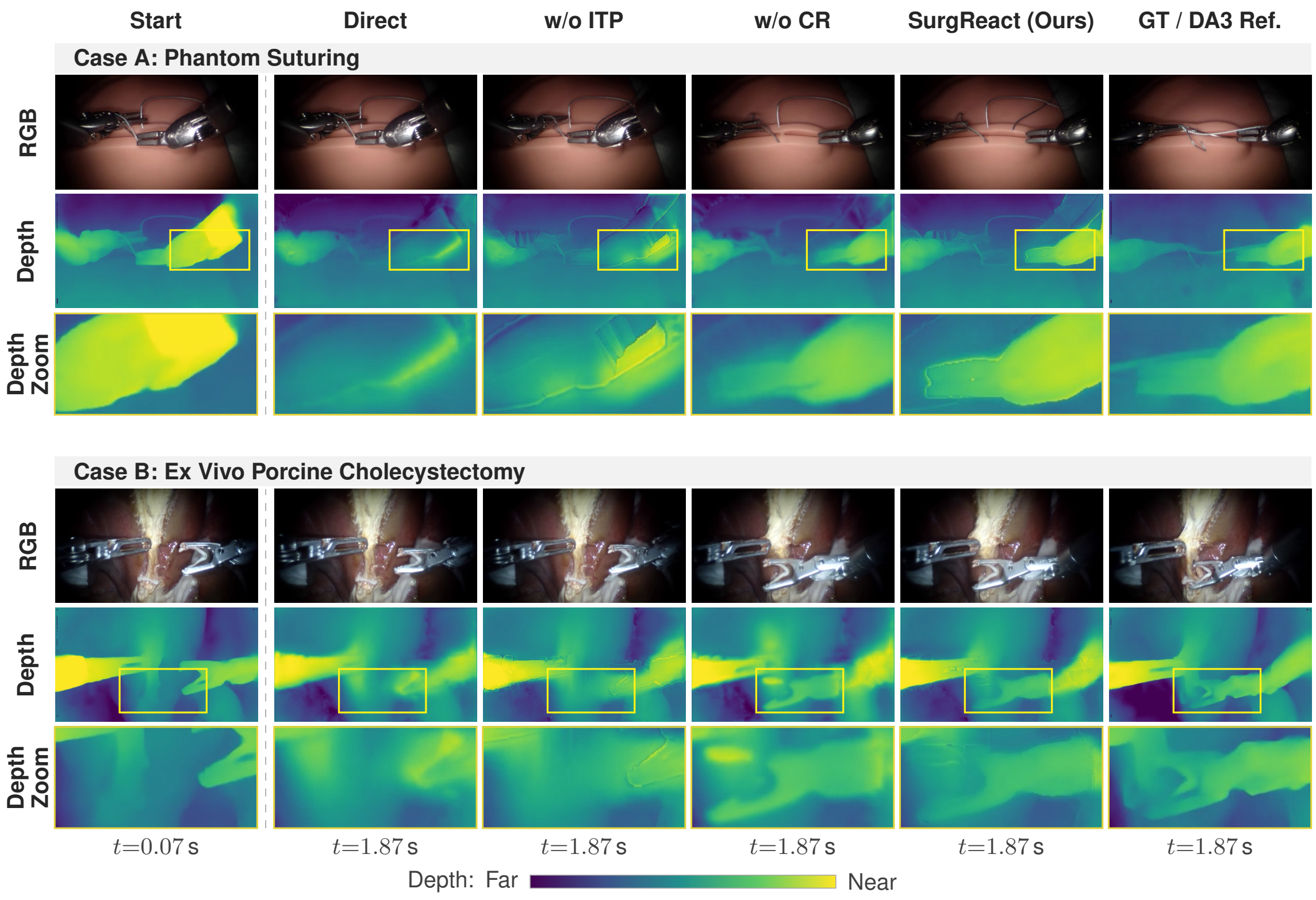

\noindent Complementing Table~2, Fig.~\ref{fig:supp_ablation_t2} compares \emph{Direct}, \emph{w/o ITP}, \emph{w/o CR}, and SurgReact across two held-out windows: Case~A on phantom suturing and Case~B on ex vivo porcine cholecystectomy. Comparing \emph{Direct}, \emph{w/o ITP}, \emph{w/o CR}, and SurgReact at $t{=}1.87$~s against the first future reference frame at $t{=}0.07$~s, Ground Truth RGB, and DA3 reference depth. All variants share the same frozen video generator and per-window depth color bounds. Without explicit decomposition in \emph{Direct} or future trajectory prediction in \emph{w/o ITP}, instruments lag near their starting poses in both RGB and depth; without residual refinement, \emph{w/o CR} roughly tracks instrument displacement but degrades local 3D surface and jaw geometry in the zoomed depth regions. SurgReact accurately captures both instrument motion and crisp, coherent 3D geometry in the interaction region.

Fig.~\ref{fig:supp_ablation_t2} illustrates how these components act together from $t{=}0.07$~s to $t{=}1.87$~s. When explicit decomposition is absent in \emph{Direct} or \emph{w/o ITP} initializes instruments from the current frame, both variants lag near their starting poses in RGB and depth: tools remain near the center during suture retraction in Case~A and fail to advance toward the central tissue during clip application in Case~B. Predicting future instrument placement through ITP therefore establishes the spatial layout of surgical interaction and improves interaction-region geometry over current-frame initialization. Meanwhile, trajectory guidance enables \emph{w/o CR} to follow coarse displacement, yet direct coordinate regression blurs and distorts local 3D jaw and surface geometry in the zoomed depth panels. By predicting future instrument geometry first and conditioning reference-based coordinate refinement on it and historical tissue geometry, SurgReact captures accurate instrument motion alongside crisp, coherent 3D interaction geometry for modeling surgical tissue reactions.

\subsection{Interactive Video Comparisons}
\label{sec:supp_qualitative}

\noindent Additional video comparisons are available on the \href{https://GenerativeIntelligence-HITSZ.github.io/SurgReact/}{Project Page}.

\section{Limitations and Future Work}
\label{sec:supp_limitations}

\noindent SurgReact relies on off-the-shelf foundation models for history preprocessing and geometric pseudo-supervision, including Depth Anything~3~\cite{ICLR2026_e4cd5012} for 3D pointmaps, SAM~3.1~\cite{ICLR2026_e0982cbc} and SurgSLOT~\cite{liu2025surgslot} for instrument segmentation, and CoTracker3~\cite{Karaev_2025_ICCV} for keypoint tracking. Under challenging endoscopic conditions such as heavy cautery smoke, prolonged occlusion, or strong specular reflections, occasional depth noise, mask identity swaps, or tracking errors introduced by these external upstream models can propagate into geometric conditioning, and running them sequentially accounts for the preprocessing latency in Table~\ref{tab:supp_runtime}. Because history perception is cleanly modularized from our instrument-conditioned geometry and video generators, directly substituting stronger or unified endoscopic perception backbones as they emerge can further improve robustness and inference speed without altering our modeling design. In addition, our downstream policy study in Sec.~4.4 focuses on offline action prediction over held-out real surgical videos; deploying SurgReact for real-time closed-loop simulation on physical surgical robotic platforms remains an exciting direction for future work.

\FloatBarrier

\end{document}